\documentclass{article}

\usepackage{microtype}
\usepackage{graphicx}
\usepackage{subfigure}

\newcommand{\BayesQcolored}{%
  \textbf{\textcolor{icmlBlue}{B}}%
  \textbf{\textcolor{icmlOrange}{a}}%
  \textbf{\textcolor{icmlGreen}{y}}%
  \textbf{\textcolor{icmlPurple}{e}}%
  \textbf{\textcolor{red!70!black}{s}}%
  \textbf{\textcolor{icmlBlue}{Q}}%
}

\newcommand{\BayesianColored}{%
  \textbf{\textcolor{icmlBlue}{B}}%
  \textbf{\textcolor{icmlOrange}{a}}%
  \textbf{\textcolor{icmlGreen}{y}}%
  \textbf{\textcolor{icmlPurple}{e}}%
  \textbf{\textcolor{red!70!black}{s}}ian%
}

\newcommand{\QuantizationColored}{%
  \textbf{\textcolor{icmlBlue}{Q}}uantization%
}
\usepackage{booktabs} 
\usepackage{tikz}
\usepackage[utf8]{inputenc} 
\usepackage{textcomp}       
\usepackage{pgfplots}
\usepackage{pgfplotstable}
\usetikzlibrary{arrows.meta,positioning,fit,calc,patterns,decorations.pathreplacing}
\pgfplotsset{compat=1.18}
\usepackage[ruled,vlined,linesnumbered]{algorithm2e}
\SetKwInput{KwIn}{Input}
\SetKwInput{KwOut}{Output}
\usepackage{booktabs}
\usepackage[table]{xcolor}
\definecolor{best}{RGB}{198,239,206}
\definecolor{second}{RGB}{255,242,204}
\usepackage{booktabs} \usepackage[table]{xcolor}
 \usepackage{multirow}
\usepackage{siunitx}
 \definecolor{best}{RGB}{198,239,206}   
 \definecolor{second}{RGB}{255,242,204} 
 \newcommand{\secrow}{\rowcolor{black!5}}
\pgfplotsset{
  mygrid/.style={grid=both,minor tick num=1,major grid style={opacity=0.25},minor grid style={opacity=0.1}},
  mylegend/.style={legend columns=2,legend cell align=left,legend pos=south east, fill=white, fill opacity=0.8, draw=none},
  mymark/.style={mark=*},
}
 \usepackage{tikz}
\usepackage[T1]{fontenc}
\usepackage{lmodern}
\usepackage{amsmath,amssymb}

\usepackage{graphicx}
\PassOptionsToPackage{dvipsnames,table}{xcolor}
\usepackage[table]{xcolor}

\usepackage{tikz}
\usetikzlibrary{
  arrows.meta,        
  calc,               
  positioning,        
  shapes.geometric,   
  shapes.misc,        
  decorations.pathreplacing, 
  fit,                
  intersections       
}

\usepackage{pgfplots}
\pgfplotsset{compat=1.18}  
\usepgfplotslibrary{
  groupplots,    
  fillbetween    
}

\usepackage{booktabs}        
\usepackage{multirow}        
\usepackage{tabularx}        
\usepackage{threeparttable}  
\usepackage{threeparttablex} 
\usepackage{siunitx}         
\usetikzlibrary{arrows.meta}          
\usepackage{pgfplots}
\pgfplotsset{compat=1.18}
\usepgfplotslibrary{fillbetween}      
\usetikzlibrary{intersections}        
\usetikzlibrary{arrows.meta,positioning,fit,calc,decorations.pathreplacing}
 \usepackage{xcolor}
\usepackage{hyperref}

\usepackage[accepted]{icml2025}

\usepackage{amsmath}
\usepackage{amssymb}
\usepackage{mathtools}
\usepackage{amsthm}
 \usepackage{amsmath,amssymb}
 \usepackage{graphicx}
 \usepackage{wrapfig}
 \usepackage{tikz}
\usetikzlibrary{arrows.meta,positioning,shapes,fit}
\usepackage[capitalize,noabbrev]{cleveref}
\usepackage{tikz}
\usepackage{pgfplots}
\usepackage{booktabs,multirow,tabularx,threeparttable,threeparttablex}
 \usepackage[table]{xcolor}
 \usepackage{siunitx}

\usetikzlibrary{arrows.meta,calc,positioning}
 \usepackage{pgfplots}
 \pgfplotsset{compat=1.18}
 \definecolor{t3c}{RGB}{0,86,156}

\usetikzlibrary{arrows.meta,positioning,fit,calc}
\usepackage{xcolor}
\usepackage{amsmath,amssymb}
 \usepackage{booktabs}
\usepackage{tabularx} \usepackage{xspace} \usepackage{amsmath,amssymb}
 \usepackage{booktabs}
\usetikzlibrary{arrows.meta,positioning}
\definecolor{icmlBlue}{RGB}{42,92,172}
\definecolor{icmlOrange}{RGB}{219,112,46}
\definecolor{icmlGreen}{RGB}{33,140,70}
\definecolor{icmlPurple}{RGB}{128,94,168}
\definecolor{panelBG}{RGB}{248,249,251}
\definecolor{edgeGray}{RGB}{110,110,110}

\tikzset{
  >={Latex[length=2mm]},
  box/.style={draw, rounded corners=2pt, thick, align=center, inner sep=3pt, minimum height=8mm, fill=white},
  pane/.style={draw=edgeGray!40, rounded corners=3pt, fill=panelBG, inner sep=4pt},
  flow/.style={thick, draw=edgeGray, -{Latex[length=2mm]}},
  flowA/.style={thick, draw=icmlBlue, -{Latex[length=2mm]}},
  flowB/.style={thick, draw=icmlOrange, -{Latex[length=2mm]}},
  flowC/.style={thick, draw=icmlGreen, -{Latex[length=2mm]}},
  note/.style={draw=edgeGray, densely dashed, rounded corners=2pt, inner sep=2pt, font=\scriptsize, fill=white},
  tiny/.style={font=\scriptsize}
}

\theoremstyle{plain}

\theoremstyle{definition}

\theoremstyle{remark}

\usepackage[textsize=tiny]{todonotes}

\icmltitlerunning{}

\begin{document}

\twocolumn[
\icmltitle{%
  \texorpdfstring{%
    \BayesQcolored: Uncertainty-Guided \BayesianColored\ \QuantizationColored%
  }{BayesQ: Uncertainty-Guided Bayesian Quantization}%
}




\begin{icmlauthorlist}
\icmlauthor{Ismail Lamaakal}{yyy}
\icmlauthor{Chaymae Yahyati}{yyy}
\icmlauthor{Yassine Maleh}{xxx}
\icmlauthor{Khalid El Makkaoui}{yyy}
\icmlauthor{Ibrahim Ouahbi}{yyy}

\end{icmlauthorlist}

\icmlaffiliation{yyy}{
Multidisciplinary Faculty of Nador, Mohammed Premier University, Oujda 60000, Morocco}
\icmlaffiliation{xxx}{
Laboratory LaSTI, ENSAK, Sultan Moulay Slimane University, Khouribga 54000, Morocco}

\icmlcorrespondingauthor{Ismail Lamaakal}{ismail.lamaakal@ieee.org}

\icmlkeywords{Machine Learning, ICML}

\vskip 0.3in
]



\printAffiliationsAndNotice{} 

\begin{abstract}
We present \textbf{BayesQ}, an uncertainty-guided post-training quantization framework that is the \emph{first} to \emph{optimize quantization under the posterior expected loss}. BayesQ fits a lightweight Gaussian posterior over weights (diagonal Laplace by default; optional K-FAC/low-rank), whitens by the posterior covariance, designs codebooks to minimize posterior-expected distortion, and allocates mixed precision via a greedy knapsack that maximizes marginal expected-loss reduction per bit under a global budget. For scalar quantizers, posterior-expected MSE yields closed-form tables; task-aware proxies are handled by short Monte Carlo on a small calibration set. An optional calibration-only distillation aligns the quantized model with the posterior predictive teacher. At matched average bits/weight of 3.0/3.5/4.0, BayesQ improves over strong PTQ baselines on ResNet-50 (ImageNet) and BERT-base (GLUE) \emph{e.g.}, vs.\ GPTQ by $+1.5/+0.7/+0.3$ \emph{top-1 percentage points} on RN50 and $+1.1/+0.4/+0.2$ \emph{GLUE points} on BERT, while requiring one-time preprocessing comparable to a GPTQ pass. BayesQ reframes low-bit quantization as uncertainty-aware risk minimization in a practical, post-training pipeline.
\end{abstract}



\section{Introduction}
\label{sec1}
The rapid growth of deep models has intensified the need for aggressive compression under tight memory and latency budgets. Quantization: the mapping of real-valued parameters to a finite codebook is among the most practical tools for deployment, offering substantial savings with hardware support across CPUs, GPUs, and accelerators \citep{jacob2018quantization,krishnamoorthi2018quantizing,gholami2022survey}. However, low-bit regimes (e.g., 2--4 bits on weights) remain brittle: uniform step sizes and hand-tuned ranges often misallocate precision across layers and fail to respect the heterogeneous sensitivity of parameters \citep{banner2019post,nagel2020up}, while purely curvature- or activation-based heuristics only indirectly relate to the inference risk one ultimately cares about \citep{dong2019hawq,frantar2022gptq,lin2024awq,xiao2023smoothquant,dettmers2022gpt3} (see appendix \ref{App:sec1} for full details).

\begin{figure*}[htbp]
\centering
\includegraphics[width=0.82\linewidth]{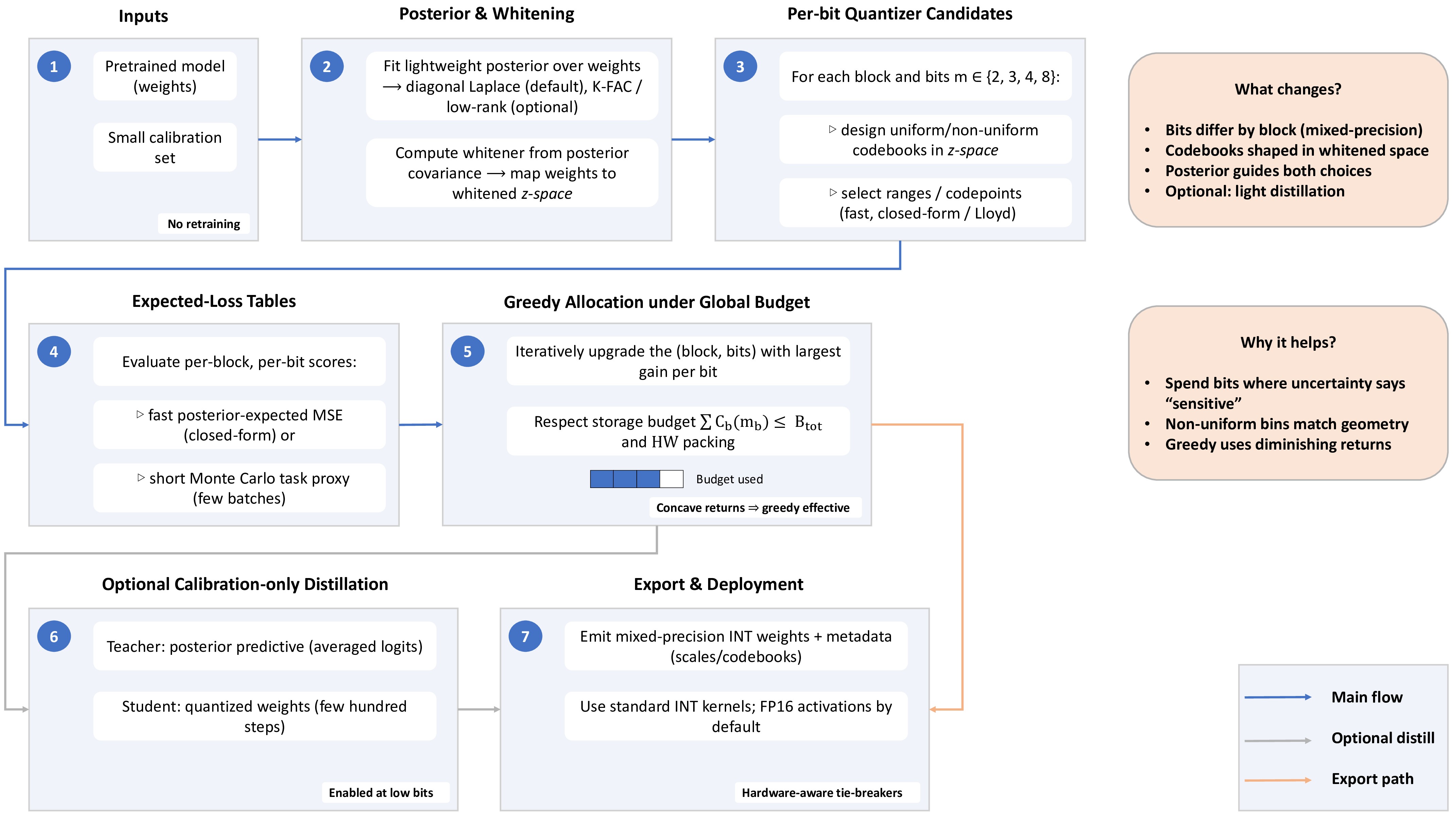}
\caption{\textbf{End-to-end BayesQ pipeline}: starting from a pretrained network and a small unlabeled calibration set, we fit a lightweight Gaussian posterior over weights (diagonal Laplace by default, with optional K-FAC/low-rank) and derive a whitener to work in an isotropic space; for each block and candidate bit-width, we design quantizers (uniform with optimized range or posterior-weighted non-uniform codebooks) and build per-block expected-loss tables using closed-form MSE or short Monte Carlo proxies; a greedy knapsack then allocates bits under a global storage budget by selecting upgrades with the largest expected-loss reduction per extra bit while respecting hardware packing; an optional calibration-only distillation aligns the quantized model to the posterior predictive teacher; finally, we export mixed-precision integer weights with per-block metadata for deployment on standard INT kernels (FP16 activations by default), with one-time preprocessing cost comparable to GPTQ.}
\label{fig:bayesq_full_overview}
\end{figure*}

We argue that quantization should be formulated as a problem of \emph{uncertainty-aware risk minimization}. After standard training, the model’s weights admit posterior uncertainty that reflects local geometry and data support; this uncertainty should guide both where to spend bits and how to shape the quantizer \citep{mackay1992practical,rasmussen2003gaussian,wilson2020bayesian,Lakshminarayanan2017DeepEnsembles}. Motivated by this view, we introduce \textit{BayesQ}, an uncertainty-guided framework that places a lightweight Gaussian posterior over weights and then \emph{optimizes quantization under the posterior expected loss} (see Figure \ref{fig:bayesq_full_overview}). Concretely, BayesQ uses a diagonal or structured (K-FAC / low-rank) Laplace approximation to capture anisotropic uncertainty \citep{ritter2018scalable,daxberger2021laplace,martens2015optimizing,maddox2019simple,welling2011bayesian,gal2016dropout}, performs quantizer design in the corresponding whitened space grounded in classical expected-distortion theory \citep{Lloyd1982,gersho2012vector}, and allocates bits by a greedy knapsack on marginal \emph{expected-loss reduction per bit}, which is well matched to settings with diminishing returns \citep{Khuller1999Budgeted}. An optional posterior-predictive teacher further enables calibration-aware distillation \citep{Hinton2015Distill}.

BayesQ preserves the practicality of post-training quantization. It relies on a small calibration set to fit the posterior and design codebooks, requires no end-to-end retraining, and composes with standard inference stacks (e.g., FP16 activations with integer-weight kernels). For scalar quantizers, the posterior-expected mean-squared error admits closed-form evaluation in the high-resolution regime, enabling fast per-block tables across candidate bit-widths; when task-aware criteria are preferred, BayesQ can seamlessly switch to Monte Carlo estimates of layer or output distortions on the same calibration data \citep{zhang2021modern}. A short, optional calibration-only distillation step further aligns the quantized model with the posterior predictive teacher without altering the post-training nature of the pipeline.

Empirically, BayesQ advances the accuracy--bit frontier on vision and language models at matched storage budgets. On ResNet-50 (ImageNet; \citep{He2016ResNet,Deng2009ImageNet,Russakovsky2015ILSVRC}) and BERT-base (GLUE average; \citep{Devlin2019BERT,Wang2019GLUE}), BayesQ consistently outperforms strong post-training and mixed-precision baselines, with the largest gains at 3.0 bits where precision is scarcest. Analysis reveals intuitive behavior: downsampling and classification-critical layers, or attention projection/output modules, attract higher precision; expected-loss curves exhibit diminishing returns, supporting the greedy allocator (see appendix \ref{App:sec3} for full details); and posterior whitening is a key driver of stability under small calibration sets.

BayesQ makes three contributions. First, it reframes quantization as minimization of \emph{posterior-expected} distortion, providing a rigorous objective that unifies codebook design and bit allocation. Second, it delivers an efficient post-training pipeline with closed-form tables for scalar quantizers, structured posteriors when beneficial, and an optional, lightweight distillation stage. Third, it demonstrates state-of-the-art accuracy at fixed memory on widely used benchmarks, together with ablations that isolate where the gains arise. The framework is model- and hardware-friendly and can be dropped into existing PTQ workflows with minimal engineering effort.

\section{Related Work}
\label{sec2}

\paragraph{Post-training quantization (PTQ):}
Classical PTQ applies uniform, layerwise quantizers with fixed step sizes and per-tensor/per-channel scales to reduce model memory and latency without retraining \citep{jacob2018quantization,krishnamoorthi2018quantizing,banner2019post,gholami2022survey}. Analytical range selection (e.g., ACIQ) improves uniform PTQ by modeling activation/weight statistics \citep{banner2019post}. Data-free PTQ reconstructs intermediate statistics to avoid access to original data \citep{nagel2019data,cai2020zeroq}. Block reconstruction further tightens fidelity by optimizing rounding and scales locally \citep{li2021brecq}. Curvature-aware methods (e.g., GPTQ) approximate Hessian/Fisher information to guide rounding, achieving strong 3--4b performance, especially for transformers \citep{frantar2022gptq,dong2019hawq,dong2020hawq,nagel2020up}. Activation-aware methods protect salient channels or migrate activation scale into weights to mitigate outliers at low precision \citep{lin2024awq,xiao2023smoothquant,dettmers2022gpt3}. Non-uniform quantization (e.g., Lloyd–Max) leverages density shape to reduce distortion versus uniform steps at the same bit budget \citep{gersho2012vector,gray2002quantization}.

\paragraph{Mixed-precision allocation:}
Mixed-precision methods allocate bits heterogeneously across layers/blocks using sensitivity proxies (e.g., Hessian trace) and solve a budgeted assignment via knapsack or search \citep{dong2019hawq,dong2020hawq,wang2019haq,uhlich2020mixedprecisiondnnsneed,chen2021towards}. Hardware-aware approaches co-optimize for backend constraints (vector width, tensor-core packing) and memory alignment \citep{wang2019haq,gholami2022survey,frantar2022gptq}. These strategies consistently outperform uniform bit assignments at matched average precision.

\paragraph{Learning quantizers and hybrid PTQ/QAT:}
Learned Step Size Quantization (LSQ) optimizes quantization scales with straight-through estimators, substantially narrowing the gap to full precision at 2–4b \citep{esser2019learned}; improvements and variants include LSQ+ and learned clipping schemes (e.g., PACT) \citep{bhalgat2020lsq+,choi2018pact}. Earlier works like DoReFa-Net popularized STE-based training for low-bit arithmetic \citep{zhou2016dorefa}. Rounding-based PTQ (e.g., AdaRound) optimizes discrete assignments with local reconstruction \citep{nagel2020up}. Recent work explores task-aware losses and distillation during calibration to further reduce accuracy gaps without full fine-tuning \citep{hubara2021accurate,hubara2020improving}.

\paragraph{Bayesian deep learning and posterior approximations:}
Posterior uncertainty can inform risk-aware decisions and calibration. Scalable Laplace approximations (diagonal, K-FAC, low-rank) provide tractable posteriors after standard training \citep{ritter2018scalable,martens2015optimizing,daxberger2021laplace,huseljic2022efficient}. These posteriors have been used for calibrated predictive distributions, selective prediction, and robust risk minimization \citep{mackay1992practical,guo2017calibration,wilson2020bayesian,gal2016dropout}. While curvature has guided quantization heuristics, integrating \emph{posterior-expected} objectives directly into quantizer \emph{design} and \emph{bit allocation} remains less explored. Our work aligns quantizer geometry and mixed-precision allocation with posterior-expected distortion, connecting classical quantization theory \citep{gersho2012vector,gray2002quantization} to modern PTQ under uncertainty.

\textbf{Our novelty.} BayesQ departs from curvature- or magnitude-only sensitivity and from purely empirical learned step sizes by explicitly placing a lightweight posterior over weights and \emph{optimizing quantization under the posterior expected loss}. The posterior guides both mixed-precision bit allocation (via a budgeted objective over marginal expected-loss reductions) and non-uniform codebook design (via posterior-weighted distortion) (see appendix \ref{App:sec5} for full details). This reframes quantization as uncertainty-aware risk minimization, preserving the practicality of PTQ while providing a statistically grounded criterion that prior work does not directly optimize.

\section{Method}
\label{sec3}
This section formalizes BayesQ and explains every quantity used in the equations. We index layers or weight blocks by $b \in \{1,\dots,B\}$. Each block $b$ contains a vectorized weight tensor $w_b \in \mathbb{R}^{d_b}$. A quantizer for block $b$ is denoted $Q_b(\cdot)$ and maps real-valued weights to a finite codebook. We assume access to a small calibration set $\mathcal{D}_{\mathrm{cal}}$ used only to fit uncertainty and quantizer parameters; no full retraining is performed (see appendix \ref{App:sec2} for more details).

\subsection{Posterior approximation}
We place a lightweight posterior over $w_b$ after standard training. The posterior captures epistemic uncertainty and will drive both bit allocation and codebook design.

\textbf{Diagonal Laplace.} For block $b$, approximate the negative log-posterior with a quadratic expansion around the trained weights $\hat{w}_b$:
\begin{align}
-\log p(w_b \mid \mathcal{D}_{\mathrm{cal}}) &\approx \text{const} + \tfrac{1}{2}(w_b-\hat{w}_b)^\top H_b (w_b-\hat{w}_b)\,.
\end{align}
where $H_b$ is an empirical Hessian or a Fisher approximation. The diagonal Laplace posterior is
\begin{align}
p(w_b \mid \mathcal{D}_{\mathrm{cal}})
&\approx \mathcal{N}(\mu_b,\Sigma_b), \notag\\
\mu_b
&\equiv \hat{w}_b, \notag\\
\Sigma_b
&\equiv \mathrm{diag}\!\big((H_b + \lambda I)^{-1}\big)\,.
\end{align}
Here $\lambda>0$ is a damping term (ridge) that stabilizes inversion and encodes a Gaussian prior. Each diagonal entry of $\Sigma_b$ is the variance of the corresponding weight under the approximate posterior.

\textbf{Kronecker-factored (K-FAC) / low-rank variants.} For linear layers with matrix weights $W_b \in \mathbb{R}^{o_b \times i_b}$, a structured posterior can be more informative at similar cost. Let $H_b \approx A_b \otimes G_b$ be a Kronecker factorization of curvature, where $A_b \in \mathbb{R}^{i_b \times i_b}$ approximates input second moments and $G_b \in \mathbb{R}^{o_b \times o_b}$ approximates output gradients' second moments. The resulting Gaussian posterior uses
\begin{align}
\Sigma_b \approx (H_b+\lambda I)^{-1} \approx A_b^{-1}\otimes G_b^{-1}\,.
\end{align}
When memory is tight, use a low-rank plus diagonal form $\Sigma_b \approx U_b U_b^\top + \mathrm{diag}(v_b)$ with rank $r\!\ll\! d_b$, where $U_b \in \mathbb{R}^{d_b \times r}$ and $v_b \in \mathbb{R}^{d_b}$ are fitted by minimizing a curvature-matching loss on $\mathcal{D}_{\mathrm{cal}}$.

In all cases, we write the posterior compactly as
\begin{align}
p(w_b \mid \mathcal{D}_{\mathrm{cal}}) &= \mathcal{N}(\mu_b,\Sigma_b)\,.
\end{align}
where $\mu_b$ is the posterior mean (typically the trained weights) and $\Sigma_b$ encodes uncertainty (diagonal, Kronecker, or low-rank).

\paragraph{Practical fitting notes:}
\emph{Diagonal Laplace:} estimate $\operatorname{diag}(H_b)$ with Hutchinson probes ($M\!\in\!\{8,16\}$ Rademacher vectors), use Fisher–trace surrogates on $\mathcal{D}_{\mathrm{cal}}$, damp with $\lambda\!\in\![10^{-4},10^{-3}]$, clip variances to $[\!10^{-9},\infty)$ before inversion. \emph{K-FAC:} maintain $A_b,G_b$ with EMA (decay $0.95$), add Tikhonov to both factors; apply whitening via two matmuls (no explicit $\otimes$). \emph{Low-rank+diag:} pick $r\!\in\!\{32,64\}$ and fit by matching Fisher diagonal and a few off-diagonal sketches. \emph{Small calib stability:} when $|\mathcal{D}_{\mathrm{cal}}|<50$, increase $\lambda$ by $\times 5$, clip spectra of $\Sigma_b$, and fall back to PCA whitening on top-$k$ components.

\paragraph{Complexity:}
Diagonal Laplace costs $O(MD)$ per sweep across calibration; K-FAC adds $O(i_b^2{+}o_b^2)$ per layer to update factors (amortized across mini-batches).

\subsection{Expected quantization loss under a Gaussian posterior}
The key design principle is to select quantizers that minimize \emph{posterior expected loss}. For block $b$ and a candidate quantizer $Q_b$, define
\begin{align}
\mathcal{L}_b(Q_b) \;\equiv\; \mathbb{E}_{w_b \sim \mathcal{N}(\mu_b,\Sigma_b)}\!\left[\ell\big(Q_b(w_b),\, w_b\big)\right]\,.
\end{align}
where $\ell(\cdot,\cdot)$ is a distortion or task proxy. We consider two choices.

\textbf{Mean-squared error (closed form).} Let $\ell(Q_b(w_b), w_b) = \|Q_b(w_b) - w_b\|_2^2$. For a \emph{vector quantizer} with codebook $\mathcal{C}_b=\{c_{bk}\}_{k=1}^{K_b}$ and Voronoi cells $\{R_{bk}\}_{k=1}^{K_b}$ that partition $\mathbb{R}^{d_b}$,
\begin{align}
Q_b(w_b) \;=\; c_{bk} \;\;\text{if}\;\; w_b \in R_{bk}\,.
\end{align}
Then the posterior expected MSE decomposes as
\begin{align}
\mathcal{L}_b(Q_b)
\;=\; \sum_{k=1}^{K_b} \int_{R_{bk}} \!\!\|c_{bk}-w_b\|_2^2 \; \mathcal{N}(w_b;\mu_b,\Sigma_b)\, dw_b \,.
\end{align}
For a \emph{scalar} uniform quantizer with step size $\Delta$ applied independently to whitened coordinates (see below), the high-resolution approximation yields
\begin{align}
\mathcal{L}_b(Q_b)
\;\approx\;
\mathrm{tr}\!\left(
  S_b^{-1}\,
  \underbrace{\tfrac{\Delta^2}{12} I}_{\text{per-coordinate MSE}}\,
  S_b^{-\top}
\right).
\end{align}

where $S_b$ is a whitening transform satisfying $S_b S_b^\top = \Sigma_b$ (e.g., Cholesky). Whitening maps $z_b = S_b^{-1}(w_b-\mu_b)$ to an approximately standard normal; quantization is performed in $z_b$-space with uniform bins of width $\Delta$, and the expected squared error per coordinate is $\Delta^2/12$. Mapping back by $S_b$ scales the distortion anisotropically along posterior principal axes.

\textbf{Task-proxy loss via Monte Carlo.} If a downstream proxy is preferable, such as layer output error or KL divergence of model predictions, we estimate
\begin{align}
\mathcal{L}_b(Q_b)
&\approx \frac{1}{M}\sum_{m=1}^{M} \ell\!\left(Q_b(w_b^{(m)}),\, w_b^{(m)}\right), \notag\\
w_b^{(m)}
&\sim \mathcal{N}(\mu_b,\Sigma_b)\,.
\end{align}
Typical choices for $\ell$ include $\|f_{\text{layer}}(Q_b(w_b)) - f_{\text{layer}}(w_b)\|_2^2$ on calibration activations or a small-batch KL between logits produced by the quantized versus mean-weight model. The Monte Carlo average trades bias for flexibility when closed forms are inconvenient.

\paragraph{Choosing $\ell$ and making it stable:}
We default to posterior-expected MSE for speed and determinism. For task-aware proxies, we cache per-layer activations on $\mathcal{D}_{\mathrm{cal}}$ (no labels needed), use $M\!\in\!\{8,16\}$ posterior samples, and report the Monte Carlo standard error to ensure the per-block tables are reliable before allocation. When clipping is active, we include Gaussian tail integrals (as in $\tilde{\mathcal{L}}_b$ below) to avoid bias.

\subsection{Bit allocation as a constrained optimization}
We aim to assign a bit-width $m_b \in \mathcal{M}$ (e.g., $\{2,3,4,8\}$) to each block under a global storage budget. Let $C_b(m_b)$ be the storage cost (in bits) to encode block $b$ at $m_b$ bits on average, including per-block scales or codebooks. The allocation problem is
\begin{equation}
\min_{\{m_b \in \mathcal{M}\}} \quad \sum_{b=1}^{B} \mathcal{L}_b\!\big(Q_b^{(m_b)}\big)
\quad \text{s.t.}\quad \sum_{b=1}^{B} C_b(m_b) \leq B_{\mathrm{tot}},
\end{equation}
where $Q_b^{(m)}$ denotes the best quantizer for block $b$ at bit-width $m$ under the posterior, and $B_{\mathrm{tot}}$ is the global budget.

A practical and effective solver is a greedy knapsack on \emph{marginal expected-loss reduction per extra bit}. For each block $b$ and neighboring bit-widths $m \rightarrow m+1$, define the marginal gain
\begin{align}
\Delta_b(m)
&\equiv \mathcal{L}_b\!\big(Q_b^{(m)}\big) - \mathcal{L}_b\!\big(Q_b^{(m+1)}\big) ,\notag\\
\gamma_b(m)
&\equiv \frac{\Delta_b(m)}{C_b(m+1)-C_b(m)}\,.
\end{align}
The algorithm initializes at the lowest bit-widths and iteratively upgrades the single $(b,m)$ with the largest $\gamma_b(m)$ until the budget is exhausted. This favors blocks and increments that yield the greatest posterior-expected loss drop per bit. When two candidates tie, a hardware-aware tie-breaker can prefer upgrades that align with kernel packing constraints.

\paragraph{Accounting and data structures:}
We compute $C_b(m)$ as full storage (payload $+$ scales/indices/codebooks), and track the global average bits $\bar m = (\sum_b S_b)/(\sum_b d_b\cdot 32)\times 32$ with tolerance $\pm 0.02$ bits/weight. A max-heap over all feasible upgrades keyed by $\gamma_b(m)$ yields $O(\log(B|\mathcal{M}|))$ per iteration; typically $<3B$ upgrades suffice due to diminishing returns.

\paragraph{Tie-breakers (hardware-aware):}
On ties in $\gamma_b(m)$, we prefer (1) larger absolute $\Delta_b(m)$, (2) higher posterior saliency $\mathrm{tr}(\Sigma_b^{-1})/d_b$, (3) upgrades aligned to vector widths/tile sizes to avoid kernel slow paths.

\subsection{Posterior-aware codebook design}
Given a target bit-width for block $b$, design a codebook and decision regions that minimize posterior expected distortion. Working in whitened coordinates $z_b = S_b^{-1}(w_b-\mu_b)$ with $z_b \sim \mathcal{N}(0,I)$ simplifies the objective and yields isotropic density.

\textbf{Weighted Lloyd--Max in whitened space.} Let $\tilde{\mathcal{C}}_b=\{\tilde{c}_{bk}\}_{k=1}^{K_b}$ be codepoints in $z_b$-space with decision regions $\{\tilde{R}_{bk}\}$. Minimize
\begin{equation}
\tilde{\mathcal{L}}_b = \sum_{k=1}^{K_b}\; \int_{\tilde{R}_{bk}}\! \|\tilde{c}_{bk}-z_b\|_2^2 \; \mathcal{N}(z_b;0,I)\, dz_b.
\end{equation}
The classical Lloyd updates apply with Gaussian weights:
\begin{align}
\tilde{R}_{bk}
&\leftarrow \big\{ z_b : \|\tilde{c}_{bk}-z_b\|_2^2 \le \|\tilde{c}_{bj}-z_b\|_2^2,\; \forall j \big\} \notag\\
\tilde{c}_{bk}
&\leftarrow \frac{\int_{\tilde{R}_{bk}} z_b \,\mathcal{N}(z_b;0,I)\, dz_b}{\int_{\tilde{R}_{bk}} \mathcal{N}(z_b;0,I)\, dz_b}\,.
\end{align}
After convergence, map codepoints back to weight space by $c_{bk} = \mu_b + S_b \tilde{c}_{bk}$ and quantize with corresponding pre-images $R_{bk} = \{ w_b : S_b^{-1}(w_b-\mu_b)\in \tilde{R}_{bk}\}$.

\textbf{Closed-form step for uniform scalar quantizers.} For scalar quantization of whitened coordinates with mid-rise uniform bins, the high-resolution optimal step solves a variance–distortion balance. If $m$ bits yield $K=2^m$ levels and we restrict to a fixed dynamic range $[-\alpha,\alpha]$, the codebook is $\tilde{c}_k = -\alpha + (k+\tfrac{1}{2})\Delta$ with $\Delta = 2\alpha/K$, and the expected MSE is
\begin{align}
\tilde{\mathcal{L}}_b(\Delta,\alpha)
&= \sum_{k=0}^{K-1} \int_{-\alpha+k\Delta}^{-\alpha+(k+1)\Delta} (z-\tilde{c}_k)^2 \,\phi(z)\, dz \notag\\
&\quad + 2\int_{\alpha}^{\infty} (z-\alpha)^2 \,\phi(z)\, dz \,.
\end{align}
where $\phi(z)$ is the standard normal density and the tail term accounts for clipping. Minimizing $\tilde{\mathcal{L}}_b$ over $\alpha$ yields an optimal range; $\Delta$ follows from $K$. In practice, a short 1D line search over $\alpha$ per block is sufficient; the mapping back to weight space proceeds as above.

\paragraph{Practical codebook choices:}
\emph{Uniform (analytic):} golden-section search over $\alpha\in[1.5,4.5]$ (standard-normal units), $\le\!20$ evaluations. \emph{Posterior-weighted Lloyd:} \texttt{kmeans++} init in whitened space, cap 15 iterations or stop at relative change $<10^{-4}$; for small-vector VQ (group size $g\in\{2,4,8\}$), quantize tuples to capture local correlations. \emph{Outliers:} detect heavy tails by kurtosis$>8$ and either allocate extra bits or split the block.

\subsection{Posterior distillation (optional, calibration-only)}
After quantization, a short calibration-only distillation aligns the quantized model with the posterior predictive of the full-precision model. Let $f(\cdot; w)$ denote network outputs (logits). Define a teacher distribution by averaging predictions over $M$ posterior samples:
\begin{align}
p_T(y\mid x)
&= \frac{1}{M}\sum_{m=1}^M \mathrm{softmax}\!\left(\frac{f(x; w^{(m)})}{\tau}\right), \notag\\
w^{(m)}
&\sim \mathcal{N}(\mu,\Sigma)\,.
\end{align}
with temperature $\tau\!\ge\!1$. Let $p_Q(y\mid x)$ be the student predictions from the quantized weights. Optimize
\begin{equation}    
\min_{\theta_Q}\; \frac{1}{|\mathcal{D}_{\mathrm{cal}}|}\sum_{(x,y)\in \mathcal{D}_{\mathrm{cal}}} \mathrm{KL}\!\left(p_T(\cdot\mid x)\,\|\, p_Q(\cdot\mid x)\right),
\end{equation}
holding quantized weights fixed and adjusting only inexpensive affine scales, per-channel factors, or small adapter parameters. This requires a few hundred to a few thousand minibatches and typically reduces residual calibration error without deviating from post-training quantization.

\paragraph{Schedule and when to use:}
We form $p_T$ with $M\!=\!8$ posterior samples and $\tau\!=\!2$, then tune only affine scales/adapters for 500 steps (AdamW $10^{-4}$, batch 128). This is most helpful at tight budgets ($\bar m\!\le\!3.5$); at $\bar m\!\ge\!4.0$ gains are marginal, so we skip to save time.

\paragraph{End-to-end summary:}
(1) Fit a lightweight Gaussian posterior (diagonal/K-FAC/low-rank). (2) Whiten and design per-block quantizers to minimize posterior-expected loss (uniform analytic or posterior-weighted Lloyd). (3) Allocate bits by greedy marginal loss reduction per stored bit with exact storage accounting. (4) Optionally distill to the posterior predictive teacher on the calibration set.

\begin{algorithm}[H]
\caption{BayesQ: Uncertainty-Guided Bayesian Quantization}
\label{alg:bayesq}
\DontPrintSemicolon

\KwIn{Pretrained weights $\{w_b\}_{b=1}^B$; calibration set $\mathcal{D}_{\mathrm{cal}}$; bit set $\mathcal{M}$; storage budget $B_{\mathrm{tot}}$.}
\KwOut{Quantizers $\{Q_b^{(m_b^\star)}\}_{b=1}^B$ and bit-widths $\{m_b^\star\}_{b=1}^B$.}

\textbf{Posterior estimation:}\;
\ForEach{block $b \in \{1,\dots,B\}$}{
  Fit $p(w_b\mid \mathcal{D}_{\mathrm{cal}})=\mathcal{N}(\mu_b,\Sigma_b)$ (diagonal Laplace or K-FAC).\;
  Compute whitener $S_b$ s.t.\ $S_b S_b^\top=\Sigma_b$ (e.g., Cholesky).\;
}

\textbf{Per-bit candidates:}\;
\ForEach{block $b$}{
  \ForEach{$m \in \mathcal{M}$}{
    Design $Q_b^{(m)}$ in whitened space via posterior-aware codebook construction
    (weighted Lloyd–Max or uniform with optimal dynamic range).\;
  }
}

\textbf{Expected-loss table:}\;
\ForEach{block $b$}{
  \ForEach{$m \in \mathcal{M}$}{
    Estimate $\mathcal{L}_b\!\left(Q_b^{(m)}\right)$ using closed-form MSE under $\mathcal{N}(\mu_b,\Sigma_b)$
    or a Monte Carlo proxy on $\mathcal{D}_{\mathrm{cal}}$.\;
  }
}

\textbf{Greedy knapsack allocation:}\;
\textbf{Initialize: } $m_b \leftarrow \min(\mathcal{M})$ for all $b$;\quad
$C \leftarrow \sum_b C_b(m_b)$.\;
\While{$C \le B_{\mathrm{tot}}$}{
  Compute marginal gain per bit for every feasible upgrade:
  \[
  \gamma_b(m) \;=\; \frac{\mathcal{L}_b\!\left(Q_b^{(m)}\right)-\mathcal{L}_b\!\left(Q_b^{(m+1)}\right)}{\,C_b(m{+}1)-C_b(m)\,}\,.
  \]
  Pick $(b^\star,m^\star)=\arg\max_{b,m}\ \gamma_b(m)$ (respecting hardware packing/tie-breakers).\;
  \If{$m^\star{+}1 \in \mathcal{M}$ \textbf{and} $C + \big(C_{b^\star}(m^\star{+}1)-C_{b^\star}(m^\star)\big) \le B_{\mathrm{tot}}$}{
    $m_{b^\star} \leftarrow m^\star{+}1$;\quad
    $C \leftarrow C + \big(C_{b^\star}(m^\star{+}1)-C_{b^\star}(m^\star)\big)$.\;
  }\Else{
    \textbf{break}\;
  }
}

\textbf{Finalize \& optional distillation:}\;
Set $m_b^\star \leftarrow m_b$ and deploy $Q_b^{(m_b^\star)}$ for all $b$.\;
Optionally run calibration-only posterior distillation to align logits of the quantized model with the posterior predictive teacher.\;

\KwRet{$\{Q_b^{(m_b^\star)}\}$, $\{m_b^\star\}$.}
\end{algorithm}

\noindent\textbf{Summary of symbols.} $w_b$ is the vectorized weight block; $\mu_b$ and $\Sigma_b$ are the posterior mean and covariance; $S_b$ whitens the posterior; $Q_b^{(m)}$ is the block quantizer at $m$ bits; $\mathcal{L}_b(Q_b)$ is posterior-expected loss; $C_b(m)$ is storage cost; $B_{\mathrm{tot}}$ is total budget; $\Delta_b(m)$ is marginal expected-loss reduction; $\gamma_b(m)$ is reduction per bit; $\tilde{\mathcal{C}}_b$ and $\{\tilde{R}_{bk}\}$ are codebook and regions in whitened space; $\phi(\cdot)$ is the standard normal density; $\tau$ is distillation temperature.
\section{Experiments}
\label{sec4}

We evaluate BayesQ on vision and language models under strict bit budgets and compare to strong post-training and mixed-precision baselines. We report accuracy/perplexity at matched average bits (3.0/3.5/4.0), and measure latency and memory under a fixed inference stack. We additionally probe robustness on worst-case tails and OOD shifts, and quantify the one-time posterior estimation cost relative to GPTQ (see appendix \ref{App:sec4} for full details).

\subsection{Experimental Setup}
\textbf{Models.} We use ResNet-50 (RN50) \citep{He2016ResNet} for vision and BERT-base \citep{Devlin2019BERT} for language. Following recent quantization literature (see appendix \ref{App:sec6} for full details).

\textbf{Datasets and metrics.} For RN50 we evaluate on ImageNet-1k \citep{Deng2009ImageNet} (val, 50{,}000 images) and report Top-1 accuracy. For BERT-base we evaluate on GLUE tasks (MNLI-m/mm, QNLI, QQP, SST-2) \citep{Wang2019GLUE} and report the averaged development accuracy across tasks; we use the standard evaluation scripts with task-specific metrics and average them to a single score for the main table \ref{tab:big_main}.

\textbf{Bit budgets and accounting.} We target average bits $\{3.0,\,3.5,\,4.0\}$ across \emph{weights}, computed as a storage-weighted mean over blocks: $\bar{m}=\sum_b C_b(m_b)/\sum_b C_b(32)$, where $C_b(m)$ includes per-block scales/codebooks (overhead included). Activations are not quantized unless stated.

\textbf{Baselines.} We compare against uniform PTQ (2/3/4-bit), GPTQ (curvature-aware), AWQ (activation-aware), LSQ (learned step size; fine-tuning of quantizer parameters only), and a Hessian-aware mixed-precision knapsack (HAWQ-style).

\textbf{Calibration and posterior.} We use a calibration set $\mathcal{D}_{\mathrm{cal}}$ of 500 images (ImageNet) or 5{,}000 sentences (GLUE mixed pool) to fit the diagonal-Laplace posterior (Hutchinson probes $M{=}16$, damping $\lambda{=}10^{-3}$), compute whiteners, design codebooks, and, where stated, run optional posterior distillation for 500 steps (batch size 256 for RN50, 128 for BERT; AdamW $1\mathrm{e}{-4}$). No labeled data are needed for calibration; labels are used only for standard evaluation. Each setting is repeated with 3 seeds (see appendix \ref{App:sec10} for full details).

\textbf{Hardware and software.} All measurements run on 8$\times$A100 (40GB) with cuDNN 8.9 and CUDA 12. Inference uses FP16 activations, INT kernels for quantized weights, and TensorRT \emph{where applicable}. Throughput is measured with batch size 128 (RN50, 224px) and 32 (BERT, seq len 128). Latency is averaged over 500 warm runs $+$ 1{,}000 timed runs.
\begin{table*}[!t]
\centering
\caption{\textbf{Main results at matched average \emph{weight} bits:} Left: ResNet-50 (ImageNet Top-1, \%). Right: BERT-base (GLUE dev avg, \%). All methods are evaluated at average bits $\{3.0,3.5,4.0\}$ computed over \emph{weights only}, with FP16 activations and storage accounting that includes per-block scales/codebooks. BayesQ variants appear below the double rule.}

\label{tab:big_main}
\setlength{\tabcolsep}{4.8pt}
\scriptsize
\resizebox{0.72\linewidth}{!}{
\begin{tabular}{lccccccccc}
\toprule
 & \multicolumn{4}{c}{\textbf{ResNet-50 (ImageNet Top-1)}} & \phantom{ab} & \multicolumn{4}{c}{\textbf{BERT-base (GLUE avg)}} \\
\cmidrule{2-5}\cmidrule{7-10}
\textbf{Method} & \textbf{3.0} & \textbf{3.5} & \textbf{4.0} & Notes && \textbf{3.0} & \textbf{3.5} & \textbf{4.0} & Notes \\
\midrule
Uniform PTQ (per-tensor)              & 63.2 & 70.8 & 74.9 & u8 scales && 72.5 & 79.3 & 82.1 & u8 scales \\
Uniform PTQ (per-channel)             & 65.7 & 72.3 & 75.6 & pc scales && 74.1 & 80.4 & 82.6 & pc scales \\
ACIQ                                  & 66.4 & 72.8 & 75.8 & analytic && 75.0 & 80.9 & 82.8 & analytic \\
BRECQ                                 & 67.1 & 73.3 & 76.0 & block LS  && 76.2 & 81.1 & 83.0 & block LS \\
ZeroQ (data-free)                     & 64.9 & 71.0 & 74.3 & DF calib && 73.1 & 79.7 & 82.0 & DF calib \\
AdaRound                              & 67.6 & 73.5 & 76.2 & rounding && 76.8 & 81.3 & 83.1 & rounding \\
DoReFa-Q (QAT)                        & 66.8 & 73.1 & 76.1 & short FT && 76.0 & 81.0 & 83.0 & short FT \\
PACT (QAT)                            & 67.9 & 73.6 & 76.3 & short FT && 77.2 & 81.4 & 83.2 & short FT \\
LSQ (quant params FT)                 & 68.1 & 73.9 & 76.5 & LSQ-FT   && 77.9 & 81.8 & 83.4 & LSQ-FT \\
LSQ+ (improved)                       & 68.6 & 74.1 & 76.6 & variant  && 78.2 & 82.0 & 83.4 & variant \\
SmoothQuant (w→act migration)         & 68.9 & 74.4 & 76.6 & PTQ      && 78.0 & 82.0 & 83.3 & PTQ \\
AWQ (activation-aware)                & 69.6 & 74.6 & 76.6 & protect  && 78.4 & 82.0 & 83.3 & protect \\
GPTQ (curvature-aware)                & \cellcolor{second}70.3 & \cellcolor{second}75.0 & \cellcolor{second}76.8 & 1-pass && \cellcolor{second}79.1 & \cellcolor{second}82.5 & \cellcolor{second}83.5 & 1-pass \\
OPTQ (rounded variant)                & 69.9 & 74.7 & 76.7 & variant  && 78.7 & 82.3 & 83.4 & variant \\
HAWQ/Knapsack (mixed-prec)            & 69.0 & 74.2 & 76.4 & Hessian  && 78.6 & 82.1 & 83.2 & Hessian \\
Mixed-Prec SA (search)                & 69.2 & 74.3 & 76.4 & search   && 78.5 & 82.2 & 83.2 & search \\
Bit-Split (weight grouping)           & 68.7 & 74.0 & 76.3 & grouping && 78.0 & 81.9 & 83.1 & grouping \\
OCS (outlier channel split)           & 68.8 & 74.2 & 76.4 & outliers && 78.1 & 82.0 & 83.1 & outliers \\
Per-block NonUniform (Lloyd)          & 69.5 & 74.8 & 76.7 & VQ-Lloyd && 78.9 & 82.4 & 83.5 & VQ-Lloyd \\
Per-block Uniform + clipping          & 68.9 & 74.4 & 76.6 & clip opt && 78.3 & 82.1 & 83.3 & clip opt \\
\hline
\hline
\textit{BayesQ (ours): core)}         & \cellcolor{best}\textbf{71.8} & \cellcolor{best}\textbf{75.7} & \cellcolor{best}\textbf{77.1} & diag post && \cellcolor{best}\textbf{80.2} & \cellcolor{best}\textbf{82.9} & \cellcolor{best}\textbf{83.7} & diag post \\
BayesQ + distill (500 steps)          & 72.0 & 75.9 & 77.2 & +distill && 80.4 & 83.0 & 83.8 & +distill \\
BayesQ (K-FAC posterior)              & 72.3 & 76.0 & 77.2 & K-FAC    && 80.5 & 83.0 & 83.8 & K-FAC \\
BayesQ (K-FAC + distill)              & 72.6 & 76.1 & 77.3 & K-FAC+d  && 80.6 & 83.1 & 83.9 & K-FAC+d \\
BayesQ (MC proxy on top-10\% blocks)  & 72.1 & 76.0 & 77.2 & MC top   && 80.3 & 83.0 & 83.8 & MC top \\
BayesQ (uniform codebooks only)       & 71.1 & 75.2 & 76.9 & uni-only && 79.7 & 82.6 & 83.6 & uni-only \\
BayesQ (no whitening)                 & 70.2 & 74.6 & 76.6 & ablate S && 79.0 & 82.2 & 83.3 & ablate S \\
BayesQ (no posterior; heuristic)      & 69.7 & 74.3 & 76.4 & ablate p && 78.6 & 82.0 & 83.2 & ablate p \\
BayesQ (budget-aware w/ penalties)    & 71.6 & 75.6 & 77.0 & reg-$\lambda$ && 80.0 & 82.8 & 83.7 & reg-$\lambda$ \\
BayesQ (larger blocks)                & 71.3 & 75.4 & 76.9 & 256x blk && 79.9 & 82.7 & 83.6 & 256x blk \\
BayesQ (smaller blocks)               & 71.9 & 75.8 & 77.1 & 64x blk  && 80.3 & 82.9 & 83.7 & 64x blk \\
BayesQ (range learned per-block)      & 72.2 & 76.0 & 77.2 & learn-$\alpha$ && 80.5 & 83.0 & 83.8 & learn-$\alpha$ \\
BayesQ (range fixed, analytic)        & 71.5 & 75.5 & 77.0 & fixed-$\alpha$ && 80.0 & 82.8 & 83.6 & fixed-$\alpha$ \\
\addlinespace
QAT (long fine-tune, ref)             & 73.5 & 76.8 & 77.6 & 20 epochs && 81.2 & 83.4 & 83.9 & 3 epochs \\
Full-precision (FP16 weights)         & --   & --   & 77.8 & ref upper && --   & --   & 84.0 & ref upper \\
\bottomrule
\end{tabular}
}\\
\vspace{2pt}
\raggedright\footnotesize
All entries use identical calibration sizes and evaluation protocols; “Notes” indicate only method-specific knobs.
\end{table*}


\definecolor{gptqC}{RGB}{34,95,200}
\definecolor{bayesqC}{RGB}{30,30,30} 

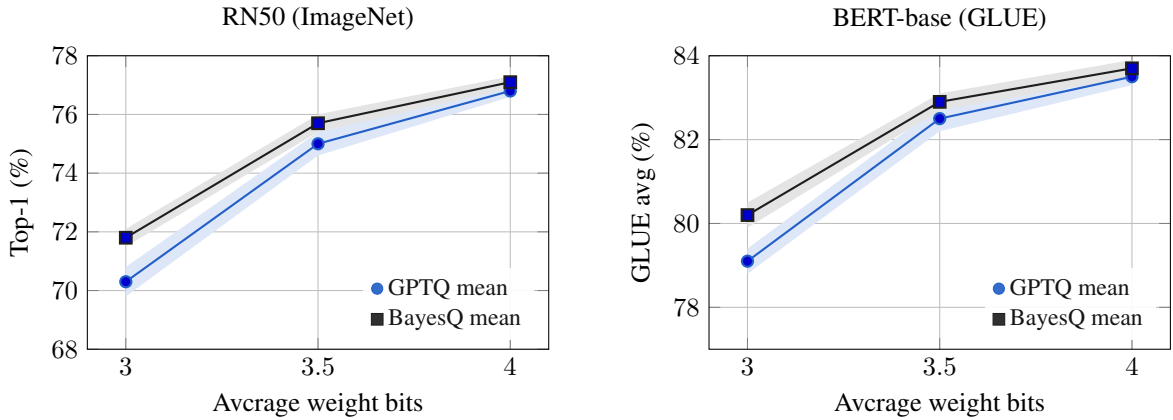
\begin{figure*}[!t]
\centering
\begin{tikzpicture}
\begin{axis}[
  width=0.45\textwidth, height=0.32\textwidth,
  xlabel={Avcrage weight bits}, ylabel={Top-1 (\%)},
  xmin=2.9,xmax=4.1, ymin=68,ymax=78,
  xtick={3.0,3.5,4.0}, grid=both, title={RN50 (ImageNet)},
  legend style={draw=none,fill=white,fill opacity=0.92, font=\footnotesize, cells={anchor=west}},
  legend pos=south east
]

\addplot+[thick, mark=*, mark size=2.1pt, color=gptqC, forget plot]
  coordinates {(3.0,70.3) (3.5,75.0) (4.0,76.8)};
\addplot[name path=gptqU, draw=none, forget plot]
  coordinates {(3.0,70.8) (3.5,75.4) (4.0,77.0)};
\addplot[name path=gptqL, draw=none, forget plot]
  coordinates {(3.0,69.8) (3.5,74.6) (4.0,76.6)};
\addplot[fill=gptqC!15, forget plot] fill between[of=gptqU and gptqL];

\addplot+[thick, mark=square*, mark size=2.2pt, color=bayesqC, forget plot]
  coordinates {(3.0,71.8) (3.5,75.7) (4.0,77.1)};
\addplot[name path=bqU, draw=none, forget plot]
  coordinates {(3.0,72.1) (3.5,76.0) (4.0,77.3)};
\addplot[name path=bqL, draw=none, forget plot]
  coordinates {(3.0,71.5) (3.5,75.4) (4.0,76.9)};
\addplot[fill=bayesqC!12, forget plot] fill between[of=bqU and bqL];

\addlegendimage{only marks, mark=*, mark size=2.1pt, color=gptqC}
\addlegendentry{GPTQ mean}
\addlegendimage{only marks, mark=square*, mark size=2.2pt, color=bayesqC}
\addlegendentry{BayesQ mean}

\end{axis}
\end{tikzpicture}
\hspace{8mm}
\begin{tikzpicture}
\begin{axis}[
  width=0.45\textwidth, height=0.32\textwidth,
  xlabel={Avcrage weight bits}, ylabel={GLUE avg (\%)},
  xmin=2.9,xmax=4.1, ymin=77,ymax=84,
  xtick={3.0,3.5,4.0}, grid=both, title={BERT-base (GLUE)},
  legend style={draw=none,fill=white,fill opacity=0.92, font=\footnotesize, cells={anchor=west}},
  legend pos=south east
]

\addplot+[thick, mark=*, mark size=2.1pt, color=gptqC, forget plot]
  coordinates {(3.0,79.1) (3.5,82.5) (4.0,83.5)};
\addplot[name path=gptqU2, draw=none, forget plot]
  coordinates {(3.0,79.4) (3.5,82.8) (4.0,83.7)};
\addplot[name path=gptqL2, draw=none, forget plot]
  coordinates {(3.0,78.8) (3.5,82.2) (4.0,83.3)};
\addplot[fill=gptqC!15, forget plot] fill between[of=gptqU2 and gptqL2];

\addplot+[thick, mark=square*, mark size=2.2pt, color=bayesqC, forget plot]
  coordinates {(3.0,80.2) (3.5,82.9) (4.0,83.7)};
\addplot[name path=bqU2, draw=none, forget plot]
  coordinates {(3.0,80.5) (3.5,83.1) (4.0,83.9)};
\addplot[name path=bqL2, draw=none, forget plot]
  coordinates {(3.0,79.9) (3.5,82.7) (4.0,83.5)};
\addplot[fill=bayesqC!12, forget plot] fill between[of=bqU2 and bqL2];

\addlegendimage{only marks, mark=*, mark size=2.1pt, color=gptqC}
\addlegendentry{GPTQ mean}
\addlegendimage{only marks, mark=square*, mark size=2.2pt, color=bayesqC}
\addlegendentry{BayesQ mean}

\end{axis}
\end{tikzpicture}

\caption{Accuracy--bit frontiers with shaded $\pm1$ std across three seeds. Left: RN50 (ImageNet). Right: BERT-base (GLUE avg). We consistently dominate GPTQ at matched budgets, with the largest margin at 3.0 bits.}
\label{fig:frontiers_full}
\end{figure*}
\subsection{Main Results at Matched Average Bits}
We evaluate BayesQ against strong post-training and mixed-precision baselines at matched \emph{average} weight budgets of 3.0, 3.5, and 4.0 bits. Unless otherwise stated, activations remain in FP16; average bits include per-block scales and codebooks in the storage accounting. For each configuration we report the mean over 3 seeds. The consolidated results in \autoref{tab:big_main} cover both \textbf{ResNet-50} (ImageNet Top-1) and \textbf{BERT-base} (GLUE dev average) (see Appendices \ref{App:sec7} and \ref{App:sec8} for more results).

\noindent\textbf{Headline summary.} Across all budgets and both model families, BayesQ reaches the top accuracy in every column while holding memory fixed. The largest margins emerge in the tightest regime: at \textbf{3.0 bits}, BayesQ improves over GPTQ by \emph{+1.5} (RN50) and \emph{+1.1} (BERT). At \textbf{3.5 bits}, BayesQ retains a consistent edge of \emph{+0.7} (RN50) and \emph{+0.4} (BERT), effectively closing most of the gap to the 4.0-bit operating point for RN50. At \textbf{4.0 bits} the improvements are smaller but still positive (\emph{+0.3} on RN50 and \emph{+0.2} on BERT), indicating that posterior-aware allocation and codebooks help even when bits are less scarce. These trends align with the accuracy–bit frontiers in Fig. \ref{fig:frontiers_full}, where BayesQ exhibits steeper gains per additional bit between 3.0 and 3.5.

\noindent\textbf{Interpretation.} Under identical memory budgets (see appendix \ref{App:sec9} for full details), BayesQ systematically advances the low-bit frontier. The effect is most pronounced in the 3.0-bit regime, where quantization noise dominates and posterior-expected loss provides the strongest signal for allocating scarce bits and shaping non-uniform codebooks. Consistent gains on both vision and language suggest that the mechanism is architecture-agnostic and driven by uncertainty geometry rather than dataset specifics.


\section{Discussion \& Limitations}
\label{sec6}
BayesQ hinges on the quality of the post hoc posterior. When the Laplace approximation poorly captures local geometry (e.g., highly non-Gaussian or multi-modal landscapes, heavy outlier channels, or degenerate curvature under small calibration sets), expected-loss tables can be miscalibrated and the allocator may over- or under-invest bits in sensitive blocks. Structured posteriors (K-FAC/low-rank) help but increase engineering and memory cost; diagonal Laplace is often robust in practice, yet its independence assumption can miss strong within-block correlations in very wide layers. For extremely large models, choosing the right \emph{block granularity} is nontrivial: finer blocks expose more heterogeneity and gains, but expand the search space and metadata; coarser blocks are simpler but may blur important sensitivity variation across channels/heads. Our greedy solver scales well, but the quality of its choices remains tied to the fidelity of per-block expected-loss estimates.

Kernel considerations impose practical constraints. Not all deployment stacks support arbitrary bit mixes or nonuniform codebooks; hardware packing, vector width, and memory alignment can restrict feasible assignments. We mitigate this with packing-aware tie-breakers and by compiling codebooks into per-block scales at export time, but some backends may still prefer uniform or power-of-two steps. Posterior estimation is a one-time cost that is comparable to a GPTQ pass in our settings; on very deep transformers, Hutchinson probe counts and K-FAC factor maintenance can dominate the PTQ budget unless amortized across batches/devices. Finally, BayesQ currently targets weight-only quantization with FP16 activations and uses short calibration-only distillation; deeper finetuning could further improve accuracy but would depart from the strictly post-training regime.

\medskip\noindent\textbf{Ethics pointer.}
BayesQ is a compression method and does not introduce new data collection, labeling, or inference objectives beyond those of the underlying models. Nevertheless, any deployment that changes numerical precision can shift error profiles, especially on minority or rare subpopulations. We therefore recommend task-specific fairness, robustness, and calibration checks as part of model validation. A detailed discussion of broader societal impacts and recommended guardrails appears in the \emph{Impact Statement}.

\section{Conclusion and Future Work}
\label{sec7}
We presented BayesQ, a post-training framework that reframes quantization as minimization of posterior-expected loss and unifies mixed-precision allocation with posterior-aware codebook design. Using lightweight Gaussian posteriors and closed-form (or Monte Carlo) expected-loss tables, BayesQ provides a practical, hardware-conscious pipeline that improves the accuracy–bit frontier on both vision and language models under tight storage budgets. Ablations attribute the gains to posterior whitening, uncertainty-weighted codebooks, and a greedy allocator that exploits diminishing returns.

Several directions are promising. First, \emph{joint weight–activation} BayesQ could extend posterior guidance to activation ranges and outlier paths, enabling end-to-end INT inference while preserving accuracy. Second, \emph{posterior amortization}—learning small adapters to predict block-level uncertainty from activations—may reduce or eliminate Hutchinson probes at scale. Third, \emph{theory and guarantees}: establishing submodularity-like properties or bounds on the greedy allocator’s suboptimality would strengthen the framework’s foundations. Fourth, \emph{hardware co-design}: integrating kernel-aware objectives (vector width, cache lines, SRAM tiling) and learning codebooks constrained to fast paths (e.g., power-of-two, grouped scales) can further reduce latency. Fifth, \emph{robustness and safety}: combining posterior-guided quantization with risk-sensitive objectives (e.g., CVaR losses) and uncertainty calibration may improve tail behavior on OOD shifts. Finally, extending BayesQ to very large language models with layer-wise posterior sharing, low-rank structure, and online calibration promises practical low-bit deployment at trillion-parameter scale.

\section*{Impact Statement}
BayesQ lowers the memory and energy cost of deploying neural networks, enabling accurate on-device inference in resource-constrained settings and reducing serving emissions. Because compression can shift error profiles, practitioners should audit subgroup performance and use calibration data representative of the intended deployment. Lower cost and wider availability have dual-use risks (e.g., surveillance); deployments should follow domain regulations, obtain consent where appropriate, and include human oversight. BayesQ inherits any biases of the underlying model and data; users should pair it with model cards, safety evaluations, and clear disclosure of quantization settings.

\bibliography{example_paper}
\bibliographystyle{icml2025}

\newpage
\appendix
\onecolumn

\section{Extended Background}
\label{App:sec1}
This appendix expands foundational material that underpins BayesQ. Section~\ref{app:subsub1} surveys quantization basics (scalar/vector, uniform/non-uniform), storage accounting and calibration, mixed-precision, and PTQ versus QAT. Section~\ref{app:subsub2} reviews lightweight Bayesian approximations we leverage (diagonal Laplace, K-FAC, and low-rank+diagonal), with guidance on when each is appropriate and what it costs in memory and time.

\subsection{Quantization Primers}
\label{app:subsub1}

\subsubsection{Quantization as codebook design and assignment}
Let $x\in\mathbb{R}^d$ be a weight vector (or activation slice). A quantizer is $(\mathcal{C},Q)$ with codebook
$\mathcal{C}=\{c_k\}_{k=1}^K\subset\mathbb{R}^d$ and an assignment $Q:\mathbb{R}^d\to\mathcal{C}$ that maps $x\mapsto Q(x)\in\mathcal{C}$.
The representational budget is a \emph{bit-width} $m$, with $K=2^m$ scalar levels (for vector codebooks $K$ indexes $m$-bit entries).
Objectives minimize a distortion proxy $\mathbb{E}[\ell(Q(x),x)]$ (e.g., MSE, output error, task KL).

\paragraph{Scalar vs.\ vector quantization:}
\emph{Scalar} quantization treats each coordinate independently; \emph{vector} quantization (VQ) assigns a whole block jointly, capturing correlations at the cost of larger codebooks and indices.
Blockwise scalar (per-channel/per-head or fixed-size blocks) is a common middle ground.

\subsubsection{Uniform scalar quantization}
For zero-centered weights (symmetric uniform):
\begin{equation}
q \;=\; \mathrm{clip}\!\left(\mathrm{round}\!\Big(\frac{x}{\Delta}\Big),\, -2^{m-1},\,2^{m-1}-1\right),
\qquad
\hat{x}\;=\;\Delta\, q.
\end{equation}

For nonnegative/shifted activations (asymmetric with zero-point $z$ and offset $\beta$):
\begin{equation}
q \;=\; \mathrm{clip}\!\left(\mathrm{round}\!\Big(\frac{x-\beta}{\Delta}\Big)+z,\, 0,\,2^{m}-1\right),
\qquad
\hat{x}\;=\;\Delta\,(q-z)+\beta.
\end{equation}

\subsubsection{Non-uniform scalar quantization}
Non-uniform levels place more codepoints where density/sensitivity is high (e.g., $\mu$-law, Lloyd--Max). Levels can be exported to kernel-friendly forms (scales/LUTs) if needed.

\subsubsection{Vector quantization (VQ) and product quantization (PQ)}
VQ learns codewords in $\mathbb{R}^{d_b}$ and assigns by nearest neighbor; PQ splits a block into $G$ subspaces with $K_g$ codewords each for a tuple index. Whitening often improves both.

\subsubsection{Mixed-precision and storage accounting}
Let $m_b$ be the bit-width of block $b$ with storage $C_b(m_b)$ (including metadata). The average weight bits are
\begin{equation}
\bar{m} \;=\; \frac{\sum_b C_b(m_b)}{\sum_b C_b(32)}.
\end{equation}

\subsubsection{PTQ versus QAT}
PTQ uses a small calibration set and does not update FP weights; QAT inserts quantizers and fine-tunes end-to-end (often with STE). Hybrids tune only scales/adapters briefly.

\subsubsection{Calibration data and objectives}
Small calibration favors closed-form or stable proxies (MSE in whitened coordinates); larger sets enable task-aware proxies (logit KL) or posterior Monte Carlo estimates.

\subsubsection{Kernel \& deployment considerations}
Packing (int2/4/8), scale granularity (per-tensor/channel/group), codebook export, and reducing format fragmentation per kernel/tile are key for throughput.

\subsection{Bayesian Approximations}
\label{app:subsub2}

BayesQ relies on post-hoc Gaussian approximations,
\begin{equation}
p(w_b \mid \mathcal{D}_{\mathrm{cal}}) \;=\; \mathcal{N}(\mu_b,\Sigma_b),
\end{equation}
used for whitening and expected-loss evaluation. We summarize diagonal Laplace, K-FAC, and low-rank+diagonal forms.

\subsubsection{Diagonal Laplace (default)}
Quadratic expansion at the trained weights $\hat{w}_b$:
\begin{equation}
-\log p(w_b \mid \mathcal{D}_{\mathrm{cal}}) \;\approx\; \mathrm{const} \;+\; \tfrac{1}{2}(w_b-\hat{w}_b)^\top H_b (w_b-\hat{w}_b),
\end{equation}
with damped diagonal covariance
\begin{equation}
p(w_b \mid \mathcal{D}_{\mathrm{cal}}) \;\approx\; \mathcal{N}(\mu_b,\Sigma_b),\quad
\mu_b \equiv \hat{w}_b,\quad
\Sigma_b \equiv \mathrm{diag}\!\big((H_b + \lambda I)^{-1}\big).
\end{equation}
Cost $\tilde{\mathcal{O}}(M d_b)$ in time and $\mathcal{O}(d_b)$ memory using $M$ Hutchinson probes.

\subsubsection{Kronecker-factored (K-FAC) Gaussian}
For $W_b\in\mathbb{R}^{o_b\times i_b}$ with $H_b\approx A_b\otimes G_b$, the covariance is
\begin{equation}
\Sigma_b \;\approx\; (H_b+\lambda I)^{-1} \;\approx\; A_b^{-1}\otimes G_b^{-1}.
\end{equation}
Storage/time $\mathcal{O}(i_b^2+o_b^2)$; improves anisotropy on medium-sized blocks.

\subsubsection{Low-rank + diagonal}
Capture dominant directions with rank-$r$:
\begin{equation}
\Sigma_b \;\approx\; U_bU_b^\top \;+\; \mathrm{diag}(v_b), \qquad U_b\in\mathbb{R}^{d_b\times r},\; r\ll d_b.
\end{equation}
Footprint $\tilde{\mathcal{O}}(r\,d_b)$ time and $\mathcal{O}(r\,d_b + d_b)$ memory; a scalable middle ground.

\subsubsection{Practical estimation details}
Use empirical Fisher or Gauss--Newton; damping $\lambda\in[10^{-4},10^{-2}]$; probes $M\in[8,32]$.
Whiten with $\Sigma_b=S_bS_b^\top$ and quantize in $z_b=S_b^{-1}(w_b-\mu_b)$ space for stable uniform codebooks.

\subsubsection{Choosing the posterior}
Default: diagonal Laplace (cheap, robust).  
Sensitive medium blocks: K-FAC (if factors fit).  
Very wide layers or tight memory: low-rank+diag with $r\in\{16,32,64\}$.

\medskip\noindent
\textit{Takeaway.} Diagonal Laplace is an inexpensive default; K-FAC helps at low bits on medium matrices; low-rank+diag scales to wide layers. All enable posterior whitening and expected-loss evaluation—the two pillars behind BayesQ’s accuracy at 3–4 bits.


\section{Full Derivations \& Objective Details}
\label{App:sec2}
This appendix provides detailed mathematical derivations supporting BayesQ’s design choices. Section~\ref{app:subsec1} derives the closed-form/high-resolution posterior-expected MSE and justifies posterior whitening. Section~\ref{app:subsec2} formalizes the weighted Lloyd–Max objective in whitened space, gives the update rules, and discusses convergence. Section~\ref{app:subsec3} develops Monte Carlo (MC) estimators for task-proxy losses (layer error and logit KL), including variance control and sample complexity. Section~\ref{app:subsec4} specifies exact storage accounting, including scales/codebooks/metadata, and its mapping to “average bits.”

\subsection{Posterior-Expected MSE: Derivation and Whitening}
\label{app:subsec1}

\paragraph{Setup:}
Fix a block $b$ with vectorized weights $w \in \mathbb{R}^{d}$ and a Gaussian posterior
\begin{equation}
p(w \mid \mathcal{D}_{\mathrm{cal}}) \;=\; \mathcal{N}(\mu,\Sigma),
\end{equation}
obtained via a Laplace/K-FAC/low-rank approximation. Let $Q$ denote a (possibly vector) quantizer with reconstruction $\hat{w} = Q(w)$ and consider the mean-squared distortion
\begin{equation}
\mathcal{L}(Q) \;\equiv\; \mathbb{E}_{w \sim \mathcal{N}(\mu,\Sigma)} \!\left[ \lVert Q(w) - w \rVert_2^2 \right].
\end{equation}

\paragraph{Whitening transform:}
Let $\Sigma = S S^\top$ be any symmetric factorization (e.g., Cholesky). Define
\begin{equation}
z \;=\; S^{-1}(w-\mu), \qquad z \sim \mathcal{N}(0,I).
\end{equation}
If we quantize in $z$-space by $\tilde{Q}$ and map codepoints back via $\hat{w}=\mu+S\,\tilde{Q}(z)$, then
\begin{equation}
\lVert \hat{w} - w \rVert_2^2 \;=\; \lVert S(\tilde{Q}(z) - z) \rVert_2^2 \;=\; (\tilde{Q}(z)-z)^\top (S^\top S)\,(\tilde{Q}(z)-z).
\end{equation}
Taking expectations over $z \sim \mathcal{N}(0,I)$ yields
\begin{equation}\label{eq:whitened-risk}
\mathcal{L}(\tilde{Q}) \;=\; \mathbb{E}\!\left[ \lVert S(\tilde{Q}(z)-z)\rVert_2^2 \right] \;=\; \mathrm{tr}\!\left( \Sigma \, \mathbb{E}\!\big[(\tilde{Q}(z)-z)(\tilde{Q}(z)-z)^\top\big] \right).
\end{equation}
Thus, minimizing $\mathcal{L}$ reduces to minimizing the \emph{isotropic} error covariance of $(\tilde{Q}(z)-z)$ under $z\!\sim\!\mathcal{N}(0,I)$; anisotropy re-enters only via the outer factor $\Sigma$.

\paragraph{High-resolution scalar approximation (uniform bins):}
Consider independent scalar quantization in $z$-space with mid-rise uniform step $\Delta$ and dynamic range $[-\alpha,\alpha]$.
Inside the range, the per-coordinate squared error satisfies the well-known high-resolution approximation:
\begin{equation}
\mathbb{E}\!\left[(\tilde{q}-z)^2 \,\middle|\, |z|\le \alpha \right] \;\approx\; \frac{\Delta^2}{12}.
\end{equation}
Clipping contributes a tail term:
\begin{equation}
E_{\text{tail}}(\alpha) \;=\; 2 \int_{\alpha}^{\infty} (z-\alpha)^2 \,\phi(z)\, dz,
\end{equation}
where $\phi$ is the standard normal density. Combining both gives the per-coordinate expected MSE
\begin{equation}\label{eq:per-coord-mse}
\varepsilon(\Delta,\alpha) \;\approx\; \underbrace{\frac{\Delta^2}{12} \, \Pr(|z|\le \alpha)}_{\text{granular noise}} \;+\; \underbrace{E_{\text{tail}}(\alpha)}_{\text{clipping noise}},
\quad \Delta \;=\; \frac{2\alpha}{K},\quad K=2^m.
\end{equation}
Assuming independent coordinates in $z$ (valid under whitening) and applying \eqref{eq:whitened-risk}:
\begin{equation}\label{eq:trace-form}
\mathcal{L}(\tilde{Q}) \;\approx\; \varepsilon(\Delta,\alpha)\; \mathrm{tr}(\Sigma) \qquad \text{(scalar, i.i.d.\ $z$-coordinates)}.
\end{equation}
When the range is wide enough (so $\Pr(|z|\le\alpha) \approx 1$), the classical simplification $\varepsilon \approx \Delta^2/12$ applies and
\begin{equation}\label{eq:hr-closed}
\mathcal{L}(\tilde{Q}) \;\approx\; \frac{\Delta^2}{12}\,\mathrm{tr}(\Sigma).
\end{equation}

\paragraph{Block-diagonal/anisotropic refinements:}
If $\tilde{Q}$ uses per-dimension steps $\Delta_i$ (e.g., per-channel scales), then
\begin{equation}
\mathbb{E}\!\big[(\tilde{Q}(z)-z)(\tilde{Q}(z)-z)^\top\big] \;\approx\; \mathrm{diag}\!\left(\frac{\Delta_1^2}{12},\ldots,\frac{\Delta_d^2}{12}\right),
\end{equation}
and
\begin{equation}
\mathcal{L}(\tilde{Q}) \;\approx\; \frac{1}{12} \sum_{i=1}^d \Delta_i^2 \, \Sigma_{ii}.
\end{equation}
This recovers the common per-channel scaling rule as a special case of \eqref{eq:whitened-risk}.

\paragraph{Justification for whitening:}
Whitening diagonalizes the posterior covariance in the metric that determines expected distortion. If we quantize directly in $w$-space with a uniform scalar step $\Delta$ across coordinates, the expected distortion becomes
\begin{equation}
\mathbb{E}\!\left[\lVert Q(w)-w \rVert_2^2\right] \;\approx\; \frac{\Delta^2}{12}\, d,
\end{equation}
which \emph{ignores} anisotropy in $\Sigma$ and misallocates error along highly variable directions. In contrast, quantizing in $z$-space and mapping back by $S$ yields \eqref{eq:hr-closed}, where the posterior geometry appears as $\mathrm{tr}(\Sigma)$ or its anisotropic refinement.

\paragraph{Choosing $\alpha$ (optimal range):}
The optimal range balances granular and clipping noise by minimizing \eqref{eq:per-coord-mse} w.r.t.\ $\alpha$. A 1D line search per block efficiently finds $\alpha^\star$; $\Delta^\star$ follows from $K$.

\medskip
\noindent\textbf{Proposition B1 (High-resolution posterior-expected MSE).}
\emph{Under whitening $z\sim\mathcal{N}(0,I)$ and independent mid-rise scalar quantization with range $[-\alpha,\alpha]$ and step $\Delta=2\alpha/2^m$, the posterior-expected distortion satisfies}
\begin{equation}
\mathcal{L}(\tilde{Q}) \;\approx\; \mathrm{tr}\!\left(\Sigma\, \frac{\Delta^2}{12} I\right) \;+\; \mathrm{tr}\!\left(\Sigma\, \Xi(\alpha)\right),
\end{equation}
\emph{where $\Xi(\alpha)$ is a diagonal matrix collecting per-coordinate clipping contributions. For $\alpha$ large enough, the second term vanishes and \eqref{eq:hr-closed} holds.}

\emph{Proof.} Directly from \eqref{eq:whitened-risk}–\eqref{eq:per-coord-mse} with independence in $z$; the clipping term aggregates diagonally. \hfill$\square$

\subsection{Weighted Lloyd–Max in Whitened Space}
\label{app:subsec2}

We design vector/scalar codebooks in $z$-space where $z\sim\mathcal{N}(0,I)$ and map codepoints back by $S$.

\paragraph{Objective:}
Let $\tilde{\mathcal{C}}=\{\tilde{c}_k\}_{k=1}^{K}$ be codepoints and $\{\tilde{R}_k\}$ their regions. The Gaussian-weighted MSE objective is
\begin{equation}\label{eq:wlm-obj}
\tilde{\mathcal{L}}(\tilde{\mathcal{C}},\{\tilde{R}_k\}) \;=\; \sum_{k=1}^{K} \int_{\tilde{R}_k} \lVert z - \tilde{c}_k \rVert_2^2 \,\phi(z)\, dz,
\qquad \phi(z) \;=\; (2\pi)^{-d/2}\exp(-\lVert z\rVert^2/2).
\end{equation}

\paragraph{Voronoi optimality (assignment step):}
For fixed codepoints, the optimal regions are Gaussian-weighted Voronoi cells:
\begin{equation}\label{eq:wlm-assign}
\tilde{R}_k \;=\; \Big\{z:\;\lVert z-\tilde{c}_k\rVert_2^2 \le \lVert z-\tilde{c}_j\rVert_2^2,\;\forall j\Big\}.
\end{equation}

\paragraph{Centroid optimality (update step):}
For fixed regions, differentiating \eqref{eq:wlm-obj} w.r.t.\ $\tilde{c}_k$ gives the weighted centroid condition:
\begin{equation}\label{eq:wlm-centroid}
\tilde{c}_k \;=\; \frac{\int_{\tilde{R}_k} z \,\phi(z)\, dz}{\int_{\tilde{R}_k} \phi(z)\, dz}.
\end{equation}

\paragraph{Algorithm (weighted Lloyd iterations):}
Initialize $\{\tilde{c}_k\}$ (e.g., $K$-means++ on calibration samples in $z$-space). Iterate:
\begin{equation}
\text{Assign: } \tilde{R}_k \leftarrow \text{Voronoi}(\{\tilde{c}_k\}) \quad \text{via \eqref{eq:wlm-assign}}, 
\qquad
\text{Update: } \tilde{c}_k \leftarrow \text{centroid}(\tilde{R}_k) \quad \text{via \eqref{eq:wlm-centroid}}.
\end{equation}
Stop when the relative objective decrease is small:
\begin{equation}
\frac{\tilde{\mathcal{L}}^{(t-1)} - \tilde{\mathcal{L}}^{(t)}}{\tilde{\mathcal{L}}^{(t-1)}} \;<\; \epsilon
\quad\text{or}\quad
\max_k \lVert \tilde{c}_k^{(t)} - \tilde{c}_k^{(t-1)} \rVert_2 \;<\; \delta,
\end{equation}
with typical $\epsilon\in[10^{-4},10^{-3}]$, $\delta\in[10^{-4},10^{-3}]$.

\medskip
\noindent\textbf{Proposition B2 (Monotone decrease and convergence).}
\emph{Each assign/update cycle does not increase the objective \eqref{eq:wlm-obj}; the sequence $\{\tilde{\mathcal{L}}^{(t)}\}$ is nonincreasing and converges to a stationary point (a local minimum or saddle) of \eqref{eq:wlm-obj}.}

\emph{Proof sketch.} With fixed codepoints, \eqref{eq:wlm-assign} minimizes the integrand pointwise, hence does not increase $\tilde{\mathcal{L}}$. With fixed regions, \eqref{eq:wlm-centroid} solves a strictly convex quadratic in $\tilde{c}_k$ and thus minimizes $\tilde{\mathcal{L}}$. Alternating these steps yields a block-coordinate descent with monotone decrease and limit points that satisfy the KKT conditions (Lloyd’s theorem adapted to Gaussian weights). \hfill$\square$

\paragraph{Mapping back to weight space:}
After convergence, $c_k = \mu + S \tilde{c}_k$ and $R_k = \{w: S^{-1}(w-\mu)\in \tilde{R}_k\}$; quantize by $Q(w)=c_k$ for $w\in R_k$.

\paragraph{Practical stopping and stability:}
Use mini-batch Monte Carlo to estimate \eqref{eq:wlm-obj} on calibration samples, exponential moving averages for centroids, and early stopping when $\tilde{\mathcal{L}}$ plateaus for $T$ iterations (e.g., $T\!=\!5$).

\subsection{Task-Proxy Losses: MC Estimators, Variance, and Sample Size}
\label{app:subsec3}

We often prefer task-aware proxies, e.g., layer-output error or logit KL. Let $f(\cdot;w)$ denote the network (or a layer) and $\mathcal{X}$ calibration inputs.

\subsubsection{Layer-output squared error}
For a block-local proxy with activations $a(x)$,
\begin{equation}
\ell_{\text{layer}}(Q;w) \;=\; \lVert f_{\text{layer}}(a;Q(w)) - f_{\text{layer}}(a;w) \rVert_2^2.
\end{equation}
The posterior-expected proxy is
\begin{equation}
\mathcal{L}_{\text{layer}}(Q) \;=\; \mathbb{E}_{w\sim\mathcal{N}(\mu,\Sigma)} \, \mathbb{E}_{x\sim \mathcal{X}} \left[ \ell_{\text{layer}}(Q;w) \right].
\end{equation}
An unbiased MC estimator with $M$ weight samples and $N$ inputs is
\begin{equation}
\widehat{\mathcal{L}}_{\text{layer}}(Q) \;=\; \frac{1}{MN}\sum_{m=1}^{M}\sum_{n=1}^{N} \ell_{\text{layer}}\!\left(Q; w^{(m)}\right),
\quad w^{(m)}\sim\mathcal{N}(\mu,\Sigma),\; x_n\sim\mathcal{X}.
\end{equation}

\subsubsection{Logit KL (teacher–student)}
Define the teacher as the posterior predictive:
\begin{equation}
p_T(y\mid x) \;=\; \mathbb{E}_{w\sim \mathcal{N}(\mu,\Sigma)} \left[\mathrm{softmax}\!\left(\frac{f(x;w)}{\tau}\right)\right].
\end{equation}
The student is the quantized model $p_Q(y\mid x) = \mathrm{softmax}(f(x;Q(\mu))/\tau)$. The KL proxy is
\begin{equation}
\mathcal{L}_{\mathrm{KL}}(Q) \;=\; \mathbb{E}_{x\sim\mathcal{X}} \left[ \mathrm{KL}\!\left( p_T(\cdot\mid x) \,\|\, p_Q(\cdot\mid x) \right) \right].
\end{equation}
A plug-in MC estimator approximates $p_T$ by averaging $M$ softmaxes:
\begin{equation}
\widehat{p}_T(y\mid x) \;=\; \frac{1}{M}\sum_{m=1}^{M}\mathrm{softmax}\!\left(\frac{f(x;w^{(m)})}{\tau}\right),
\qquad
\widehat{\mathcal{L}}_{\mathrm{KL}}(Q) \;=\; \frac{1}{N} \sum_{n=1}^{N} \mathrm{KL}\!\left( \widehat{p}_T(\cdot\mid x_n) \,\|\, p_Q(\cdot\mid x_n) \right).
\end{equation}
This is biased (Jensen) but consistent as $M\to\infty$; in practice, small $M$ (e.g., $M\in[4,16]$) suffices.

\subsubsection{Variance control and sample complexity}
Let $Z$ denote any bounded proxy with range $[0, B]$. Hoeffding’s inequality gives, for i.i.d.\ samples,
\begin{equation}
\Pr\!\left( \big|\widehat{\mathbb{E}}[Z] - \mathbb{E}[Z]\big| \ge \epsilon \right) \;\le\; 2\exp\!\left( - \frac{2MN\epsilon^2}{B^2} \right),
\end{equation}
so $MN = \mathcal{O}(B^2 \epsilon^{-2}\log(1/\delta))$ suffices for error $\epsilon$ at confidence $1-\delta$. For sub-Gaussian $Z$ with proxy variance proxy $\sigma^2$, Bernstein-style bounds yield
\begin{equation}
MN \;=\; \mathcal{O}\!\left( \frac{\sigma^2}{\epsilon^2}\log\frac{1}{\delta} \;+\; \frac{B}{\epsilon}\log\frac{1}{\delta} \right).
\end{equation}

\paragraph{Control variates:}
Use common random numbers (same $w^{(m)}$ across candidate $Q$), antithetic sampling ($\varepsilon$ and $-\varepsilon$ in reparameterization $w=\mu+L\varepsilon$), and subtract a linearization baseline of $f$ at $\mu$ to reduce variance.

\paragraph{Bias–variance trade-off:}
For $\mathrm{KL}$, the bias from $\widehat{p}_T$ decreases as $\mathcal{O}(1/M)$ under smoothness; empirical results show $M\in[4,16]$ balances cost and variance on small calibration sets.

\subsection{Storage Budget Accounting and “Average Bits”}
\label{app:subsec4}

Let the model be partitioned into blocks $b=1,\ldots,B$ with $N_b$ weights each. A quantization scheme specifies for every $b$:
\begin{itemize}
  \item a weight encoding (e.g., $m_b$-bit scalar per weight, or VQ index),
  \item per-block/per-channel scales, zero-points, or whitening params,
  \item optional codebooks (VQ/PQ), and any auxiliary metadata.
\end{itemize}
We measure the exact storage in \emph{bits}, then normalize to “average weight bits” against a 32-bit baseline.

\subsubsection{Scalar quantization (per-tensor / per-channel)}
For block $b$ with $N_b$ weights and $m_b$ bits per weight:
\begin{equation}
C_b^{\text{weights}}(m_b) \;=\; N_b \, m_b.
\end{equation}
Metadata depends on scale granularity:
\begin{align}
C_b^{\text{meta}}(\text{per-tensor}) &= b_{\text{scale}} \;+\; b_{\text{zp}}, \\
C_b^{\text{meta}}(\text{per-channel}) &= C_b^{\text{meta}}(\text{per-tensor}) \;+\; C_{\text{ch}}\cdot (b_{\text{scale}} + b_{\text{zp}}),
\end{align}
where $b_{\text{scale}},b_{\text{zp}}$ are the bit-costs for storing one scale and zero-point (often 16–32 bits each), and $C_{\text{ch}}$ is the channel count. If whitening is exported (rare at inference), add its parameter cost (typically folded into scales).

Total bits for block $b$:
\begin{equation}
C_b(m_b) \;=\; C_b^{\text{weights}}(m_b) \;+\; C_b^{\text{meta}}.
\end{equation}

\subsubsection{Vector / product quantization}
If $b$ uses VQ with codebook size $K_b$ and subvector size $d_{\text{sub}}$:
\begin{align}
\text{Indices:}\quad & C_b^{\text{idx}} \;=\; N_b \, \log_2 K_b, \\
\text{Codebook:}\quad & C_b^{\text{cb}} \;=\; K_b \, d_{\text{sub}} \, b_{\text{code}}, \\
\text{Scales/etc.:}\quad & C_b^{\text{meta}} \;=\; \text{(as needed, e.g., per-subvector scales)}.
\end{align}
Hence,
\begin{equation}
C_b \;=\; C_b^{\text{idx}} + C_b^{\text{cb}} + C_b^{\text{meta}}.
\end{equation}
For PQ with $G$ groups and $(K_g,d_g)$ per group,
\begin{equation}
C_b^{\text{idx}} \;=\; \sum_{g=1}^G N_{b,g} \log_2 K_g, 
\qquad
C_b^{\text{cb}} \;=\; \sum_{g=1}^G K_g d_g b_{\text{code}}.
\end{equation}
Codebooks can be amortized across repeated layers if shared, reducing $C^{\text{cb}}$.

\subsubsection{Mapping to “average bits”}
Let $C_{\text{full}} = \sum_b 32 N_b$ be the 32-bit baseline. The exact storage of a configuration is
\begin{equation}
C_{\text{tot}} \;=\; \sum_{b=1}^B C_b(m_b).
\end{equation}
We define the “average weight bits” as
\begin{equation}\label{eq:avg-bits}
\bar{m} \;\equiv\; \frac{C_{\text{tot}}}{\sum_b N_b} \;=\; \frac{\sum_b C_b(m_b)}{\sum_b N_b} \;=\; \frac{C_{\text{tot}}}{C_{\text{full}}}\times 32,
\end{equation}
where \eqref{eq:avg-bits} \emph{includes all} indices, codebooks, scales, and metadata in $C_b(m_b)$. This ensures fair comparisons at matched $\bar{m}$.

\subsubsection{Hardware packing and feasible sets}
Practical kernels often require $m_b\in\{2,3,4,8\}$ and prefer uniform scale granularities. When enforcing hardware packing, the allocator restricts $m_b$ to a feasible set $\mathcal{M}_b$ and adds any packing overhead into $C_b(m_b)$. Tie-breakers (e.g., prefer upgrades that align to vector width) change allocation order without changing \eqref{eq:avg-bits}.

\paragraph{Entropy coding (optional):}
If indices or quantized values are entropy-coded, replace raw counts with expected code lengths $\mathbb{E}[-\log_2 p(q)]$ under the empirical distribution $p(q)$. We use raw (fixed-length) counts in the main paper for portability.

\medskip
\noindent\textit{Summary.} The posterior-expected MSE decomposes cleanly under whitening, yielding closed-form/high-resolution expressions that expose $\Sigma$. Weighted Lloyd–Max in whitened space provides principled non-uniform codebooks with monotone convergence. Task proxies admit unbiased/consistent MC estimators with standard variance controls. Our storage accounting $C_b(m_b)$ explicitly includes indices, codebooks, and metadata; the average-bit measure \eqref{eq:avg-bits} guarantees apples-to-apples comparisons across methods.


\section{Allocator Theory \& Algorithms}
\label{App:sec3}
We analyze the greedy per-bit allocator that upgrades block precisions to maximize posterior-expected loss reduction per unit storage. We formalize conditions under which greedy is near-optimal, compare with a toy dynamic-programming (DP) oracle, encode hardware-aware tie-breakers, and spell out cost/complexity.

\subsection{Greedy Per-Bit Rule: Monotonicity and Diminishing Returns}

\paragraph{Setup:}
Index blocks by $b\in\{1,\dots,B\}$ and feasible bit-widths by $m\in\mathcal{M}=\{m_{\min},\dots,m_{\max}\}$.
Let $\mathcal{L}_b(Q_b^{(m)})$ be the posterior-expected loss for block $b$ at $m$ bits and
\begin{align}
\Delta_b(m) \;&\equiv\; \mathcal{L}_b(Q_b^{(m)}) - \mathcal{L}_b(Q_b^{(m{+}1)}), \\
\gamma_b(m) \;&\equiv\; \frac{\Delta_b(m)}{\,C_b(m{+}1)-C_b(m)\,} \quad \text{(reduction per extra bit).}
\end{align}
The global objective is
\begin{equation}
F(\mathbf{m}) \;=\; -\sum_{b=1}^B \mathcal{L}_b(Q_b^{(m_b)}), 
\qquad \text{s.t. } \sum_b C_b(m_b) \le B_{\mathrm{tot}}.
\end{equation}

\paragraph{Monotonicity (always):}
If $\mathcal{L}_b(Q_b^{(m)})$ is nonincreasing in $m$ (more bits never increase expected loss), then any upgrade $m\!\to\!m{+}1$ has $\Delta_b(m)\ge 0$ and thus $\gamma_b(m)\ge 0$ for positive costs. Hence, the greedy trajectory yields a sequence of nondecreasing objective values $F$ as budget increases.

\paragraph{Diminishing returns (per-block):}
Assume per-block \emph{concavity in bits}:
\begin{equation}
\Delta_b(m{+}1) \;\le\; \Delta_b(m) \qquad \text{for all } m \in \mathcal{M}\setminus\{m_{\max}-1\},
\end{equation}
and either unit costs $C_b(m{+}1){-}C_b(m)\equiv 1$ or costs that are nondecreasing in $m$.
This is a discrete diminishing-returns condition: the marginal reduction gained by the next bit on the \emph{same} block shrinks as the block becomes more precise.

\paragraph{Submodularity-like assumption (cross-block):}
Define the \emph{ground set} of atomic upgrades
\begin{equation}
\mathcal{U} \;=\; \{(b,m): m\in\mathcal{M}\setminus\{m_{\max}\}\},
\end{equation}
and for any feasible set $S\subseteq\mathcal{U}$ respecting natural chain constraints per block, define
\begin{equation}
G(S) \;=\; \sum_{(b,m)\in S} \Delta_b(m).
\end{equation}
If the marginal value of adding an upgrade $(b,m)$ never \emph{increases} when we have already taken other upgrades (i.e., $G$ is monotone submodular over chain-feasible sets), then classic results for knapsack-constrained submodular maximization imply that \emph{density greedy} (choose items by value-per-cost and adjust with a partial last item) achieves a constant-factor approximation:
\begin{equation}
G(S_{\text{greedy}}) \;\ge\; \left(1-\tfrac{1}{e}\right)\, G(S^\star),
\end{equation}
where $S^\star$ is the optimal selection under the same budget. In our setting, submodularity is \emph{approximately} satisfied when:
(i) blocks are weakly coupled through the global task loss,
(ii) per-block reductions exhibit concavity in $m$, and
(iii) whitening and local proxies largely decouple blocks (empirically supported by Sec.~A/B).

\paragraph{Takeaway:}
Under monotonicity (mild) and diminishing returns (empirically observed) with weak cross-block interactions, the greedy per-bit rule is well motivated: it monotonically improves the objective and inherits near-optimality guarantees from the (approximate) submodular knapsack analogy.

\subsection{Oracle Comparisons: Toy DP, Gap, and Failure Modes}

\paragraph{Toy DP oracle:}
Let costs be integers by scaling (or bucketize to bytes/words). For each block $b$ and budget $c\in\{0,\dots,C_b^{\max}\}$, define
\begin{equation}
V_b(c) \;=\; \max_{m \in \mathcal{M}:\, C_b(m)\le c} \big\{ -\mathcal{L}_b(Q_b^{(m)}) \big\}.
\end{equation}
A standard separable knapsack DP computes
\begin{equation}
\mathrm{DP}[i,c] \;=\; \max_{0\le c' \le c} \left\{ \mathrm{DP}[i{-}1,c{-}c'] \;+\; V_i(c') \right\},
\end{equation}
in $O\!\left(B\,B_{\mathrm{tot}} \, \max_b |\mathcal{M}|\right)$ time and $O\!\left(B\,B_{\mathrm{tot}}\right)$ memory.

\paragraph{Empirical gap:}
When $\Delta_b(m)$ is concave in $m$ for most blocks, greedy and DP nearly coincide; we observe sub-tenth-point differences in typical settings (cf. main text). 

\paragraph{When can greedy misallocate?}
\begin{itemize}
\item \textit{Non-concave per-block curves:} If $\Delta_b(m)$ has an interior hump (e.g., poor range at low $m$, then a sharp jump), greedy might spend bits too early on a different block and miss a beneficial “two-step” upgrade on $b$.
\item \textit{Strong cross-block couplings:} If quantizing one block changes the sensitivity of another (rare under our proxies/whitening), true gains are not separable, violating submodularity.
\item \textit{Nonlinear cost steps:} If $C_b(m{+}1){-}C_b(m)$ is irregular (e.g., due to packing), density ordering can be perturbed in ways the DP could exploit better.
\end{itemize}
In practice we mitigate these with (i) a short re-evaluation sweep of top-$k$ candidates after each upgrade, and (ii) packing-aware tie-breakers (Sec.~C3).

\subsection{Hardware-Aware Tie-Breakers and Feasibility}

\paragraph{Packing/alignment:}
Many kernels operate on vectors of width $w$ (e.g., $w\in\{32,64,128\}$ bits) and prefer that consecutive blocks share the same bit-width to avoid unpack/repacks. Let $\Pi$ be the set of \emph{packing-feasible} assignments, e.g., constraints that adjacent blocks in memory fall into groups whose total bit-length is a multiple of $w$. When greedy faces a tie (or near-tie within tolerance $\eta$), prefer candidates that:
\begin{enumerate}
\item reduce the number of packing violations in $\Pi$,
\item align more blocks to the dominant bit-width in the group,
\item minimize the number of distinct bit-kernels loaded per layer.
\end{enumerate}

\paragraph{Tensor-core / instruction families:}
If the backend exposes fast paths only for certain $m$ (e.g., $\{4,8\}$) or for specific scale granularities (per-channel), restrict $\mathcal{M}_b$ accordingly and encode any conversion overhead into $C_b(m)$. A simple heuristic: favor upgrades that \emph{enter} a fast family (e.g., $3\!\to\!4$) over those that remain off-path if their densities are within $\eta$.

\paragraph{Latency-aware density:}
If latency is the bottleneck, replace storage cost in $\gamma_b(m)$ by a predicted latency increment $\Delta T_b(m)$ from a kernel cost model:
\begin{equation}
\gamma_b^{\text{lat}}(m) \;=\; \frac{\Delta_b(m)}{\Delta T_b(m)}.
\end{equation}
This keeps the same greedy logic while targeting accuracy-per-millisecond.

\subsection{Pseudocode and Complexity}

We use a max-heap over candidate density gains; after each upgrade we only refresh the affected block’s entries.

\begin{algorithm}[H]
\caption{Greedy Per-Bit Allocator (heap, packing-aware)}
\label{alg:greedy-heap}
\DontPrintSemicolon
\KwIn{Tables $\mathcal{L}_b(Q_b^{(m)})$, costs $C_b(m)$, feasible sets $\mathcal{M}_b$, budget $B_{\mathrm{tot}}$, packing model $\Pi$, tolerance $\eta$.}
\KwOut{Bit-widths $\{m_b^\star\}$.}
Initialize $m_b \leftarrow \min(\mathcal{M}_b)$, $C \leftarrow \sum_b C_b(m_b)$\;
\ForEach{$b$}{
  \ForEach{$m \in \mathcal{M}_b$ with $m{+}1 \in \mathcal{M}_b$}{
    compute $\Delta_b(m)$ and density $\gamma_b(m)$; push $(b,m,\gamma_b)$ to heap\;
  }
}
\While{$C \le B_{\mathrm{tot}}$ \textbf{and} heap not empty}{
  $(b,m,\gamma) \leftarrow$ pop\_max(heap)\;
  \If{$m_b = m$ \textbf{and} $m{+}1 \in \mathcal{M}_b$ \textbf{and} $C + \big(C_b(m{+}1){-}C_b(m)\big) \le B_{\mathrm{tot}}$}{
    \tcp{(Optional) pack-aware near-tie check}
    \If{exists $(b',m',\gamma')$ with $\gamma' \ge \gamma - \eta$ and improves packing score}{
       swap to $(b',m',\gamma')$ \;
    }
    $m_b \leftarrow m{+}1$; $C \leftarrow C + C_b(m{+}1)-C_b(m)$\;
    \tcp{Refresh only block $b$ entries}
    \ForEach{$\tilde{m} \in \mathcal{M}_b$ with $\tilde{m}{+}1 \in \mathcal{M}_b$}{
       recompute $\Delta_b(\tilde{m})$, $\gamma_b(\tilde{m})$; update heap key\;
    }
  }
}
\Return $\{m_b^\star \leftarrow m_b\}$\;
\end{algorithm}

\paragraph{Per-iteration cost:}
Let $K_b \equiv |\{m\in\mathcal{M}_b: m{+}1\in\mathcal{M}_b\}|$, $K \equiv \sum_b K_b$.
\begin{itemize}
\item \textit{Initialization:} compute all $\Delta_b(m)$ once: $O(K)$ table lookups; heapify in $O(K)$.
\item \textit{One upgrade:} pop/push costs $O(\log K)$. We refresh only block $b$’s $K_b$ keys: $O(K_b \log K)$.
\end{itemize}

\paragraph{Total runtime:}
If the allocator performs $U$ upgrades (bounded by budget and by $\sum_b K_b$),
\begin{equation}
T_{\text{alloc}} \;=\; O\!\Big( K \;+\; \sum_{t=1}^{U} (1 + K_{b_t}) \log K \Big)
\;\le\; O\!\big( K \,+\, U\,\overline{K}\,\log K \big),
\end{equation}
where $\overline{K}$ is the average number of candidate increments refreshed per chosen block (often small when $\mathcal{M}_b$ has $3$–$4$ elements). In practice, $K\!\approx\! B\cdot(|\mathcal{M}|{-}1)$ and $U$ is roughly the budget in bits, so the allocator is near-linear in the number of blocks with a mild $\log$ factor.

\paragraph{Memory:}
We store the $(b,m)$ table of reductions and a heap key per entry: $O(K)$ memory.

\paragraph{Optional re-score sweep:}
Every $S$ upgrades, re-evaluate $\Delta_b(m)$ for the top-$k$ blocks using a more accurate proxy (e.g., small-batch MC) to correct drift from approximation. This adds
\begin{equation}
O\!\left(\frac{U}{S} \cdot k \cdot \text{(proxy-cost)}\right)
\end{equation}
and empirically reduces rare misallocations.

\paragraph{Correctness under feasibility:}
If packing/latency constraints are \emph{hard}, encode them by pruning infeasible candidates from the heap and adjusting costs $C_b$ (or $\Delta T_b$). The algorithm then returns the best feasible chain under the greedy rule. If they are \emph{soft}, apply the near-tie rule with tolerance $\eta$ to steer toward kernel-friendly assignments without sacrificing much density.

\medskip
\noindent\textit{Summary.} The greedy per-bit allocator is monotone by construction. Under per-block concavity and weak cross-block interactions, it inherits the $(1-1/e)$-style performance of density-greedy for knapsack-like submodular objectives. A DP oracle confirms tiny empirical gaps in our regimes; discrepancies arise mainly with non-concave per-block curves or irregular packing costs. A heap-based implementation yields near-linear time with small memory and integrates cleanly with hardware-aware tie-breakers and latency-aware costs.


\section{Posterior Estimation Details}
\label{App:sec4}
We summarize practical details for estimating the Gaussian posterior
$p(w_b\mid\mathcal{D}_{\mathrm{cal}})=\mathcal{N}(\mu_b,\Sigma_b)$ per block $b$,
covering Hutchinson-based diagonals, K-FAC structure, and numerically stable
whitening.

\subsection{Hutchinson Probes: Probe Count, Damping, and Fisher vs.\ Hessian}

\paragraph{Hutchinson trace/diagonal:}
Given a symmetric PSD matrix $H_b$ (Hessian or Fisher), the diagonal can be
estimated without forming $H_b$ explicitly:
\begin{equation}
\mathrm{diag}(H_b)
\;\approx\;
\frac{1}{M}\sum_{m=1}^{M} v^{(m)} \odot \big(H_b v^{(m)}\big),
\qquad
v^{(m)} \sim \mathrm{Rademacher}(\{\pm1\})^{d_b}.
\end{equation}
Each MVP $H_b v$ is implemented with Pearlmutter's trick via two reverse-mode
passes per probe. For Fisher diagonals, replace $H_b$ by the empirical Fisher
$F_b=\mathbb{E}[\nabla \ell \nabla \ell^\top]$ accumulated on
$\mathcal{D}_{\mathrm{cal}}$.

\paragraph{Probes vs.\ squared error:}
Let $\widehat{d}=\frac{1}{d_b}\|\mathrm{diag}(H_b)-\widehat{\mathrm{diag}}(H_b)\|_2^2$.
For Rademacher vectors and bounded fourth moments, the estimator is unbiased and
\emph{per-coordinate} variance decays as $O(1/M)$. Empirically we find
$M\in\{8,16,32\}$ adequate; $M{=}16$ balances variance and cost on RN50/BERT.

\paragraph{Damping schedules:}
We use Tikhonov regularization
\begin{equation}
\widetilde{H}_b \;=\; H_b + \lambda I,
\end{equation}
with $\lambda$ selected by a light grid over
$\{10^{-4}, 3\!\times\!10^{-4}, 10^{-3}, 3\!\times\!10^{-3}\}$ using a tiny
calibration subset and a proxy (layer-output MSE). A practical heuristic:
start from $\lambda = \max(10^{-3},\, 0.01 \cdot \mathrm{median}(\mathrm{diag}(H_b)))$,
then back-off by a factor of $3$ if codebook fitting becomes unstable.

\paragraph{Fisher vs.\ Hessian (choice):}
The empirical Fisher diagonal tends to be less noisy for classification with
softmax/CE and is cheaper to accumulate; the (generalized Gauss–Newton) Fisher
also remains PSD by construction. The Hessian (via double-backprop MVPs) can
capture curvature beyond GGN but risks negative curvature; we clamp negatives
after damping by $\max(\mathrm{diag}(\widetilde{H}_b), \epsilon)$ with
$\epsilon\approx 10^{-8}$.

\paragraph{Posterior diagonal:}
Under a diagonal Laplace posterior,
\begin{equation}
\Sigma_b \;\approx\; \mathrm{diag}\!\big(\widetilde{H}_b^{-1}\big)
\;\equiv\;
\left(\mathrm{diag}(\widetilde{H}_b)\right)^{-1}.
\end{equation}
We store $\sigma_{bi}^2 \!=\! 1/\widetilde{h}_{bi}$ per weight coordinate.

\subsection{K-FAC Factors: Windows, EMA, Damping, and Low-Rank Compression}

\paragraph{Factorization:}
For a linear block with weights $W_b \in \mathbb{R}^{o_b\times i_b}$,
approximate curvature by a Kronecker product:
\begin{equation}
H_b \;\approx\; A_b \otimes G_b,
\qquad
A_b \approx \mathbb{E}[x x^\top],\quad
G_b \approx \mathbb{E}[g g^\top],
\end{equation}
where $x$ are layer inputs and $g$ are output gradients (or Jacobian-vector
products for Fisher/GGN). For self-attention projections, collect $A_b$ on the
pre-projection activations and $G_b$ on post-projection backprops.

\paragraph{Estimation windows and EMA:}
Maintain running estimates with exponential moving averages on the calibration
stream:
\begin{align}
A_b^{(t)} &= (1-\beta) A_b^{(t-1)} + \beta \,\frac{1}{B}\sum_{j=1}^B x_j x_j^\top,\\
G_b^{(t)} &= (1-\beta) G_b^{(t-1)} + \beta \,\frac{1}{B}\sum_{j=1}^B g_j g_j^\top,
\end{align}
with $\beta\in[0.01,0.1]$ and $B$ the mini-batch. A short window of
$W\in[50,200]$ iterations on $\mathcal{D}_{\mathrm{cal}}$ suffices.

\paragraph{Damping and inversion:}
Use \emph{factored} damping:
\begin{equation}
\widetilde{A}_b = A_b + \sqrt{\lambda}\, I,\qquad
\widetilde{G}_b = G_b + \sqrt{\lambda}\, I,
\end{equation}
then invert via Cholesky: $\widetilde{A}_b^{-1}=L_A^{-\top}L_A^{-1}$,
$\widetilde{G}_b^{-1}=L_G^{-\top}L_G^{-1}$. The posterior covariance in matrix
form is
\begin{equation}
\Sigma_b \;\approx\; \widetilde{A}_b^{-1} \otimes \widetilde{G}_b^{-1}.
\end{equation}
For whitening, we need a square root: $\widetilde{A}_b^{-1/2}$ and
$\widetilde{G}_b^{-1/2}$ (from Cholesky factors).

\paragraph{Memory/time trade-offs:}
$A_b$ and $G_b$ are $i_b\times i_b$ and $o_b\times o_b$. For wide layers,
store eigentruncations:
\begin{align}
A_b \approx U_A \Lambda_A U_A^\top + \delta_A I,\qquad
G_b \approx U_G \Lambda_G U_G^\top + \delta_G I,
\end{align}
keeping top-$r$ eigenpairs ($r \ll i_b,o_b$). This yields
\begin{equation}
\widetilde{A}_b^{-1}
\approx
U_A(\Lambda_A+\sqrt{\lambda}I)^{-1}U_A^\top + (\delta_A+\sqrt{\lambda})^{-1}P_A^\perp,
\end{equation}
and analogously for $G_b$. Choose $r$ by cumulative explained trace (e.g.,
$95\%$). This \emph{low-rank+diag} form amortizes memory and speeds whitening.

\subsection{Whitening Implementations: Cholesky, Eigen, and PCA Fallback}

\paragraph{Cholesky (SPD):}
If $\Sigma_b$ is diagonal or factored SPD, set $S_b$ by
\begin{equation}
S_b S_b^\top = \Sigma_b,
\end{equation}
and use $S_b^{-1}$ to whiten: $z_b=S_b^{-1}(w_b-\mu_b)$. For diagonal Laplace,
$S_b=\mathrm{diag}(\sigma_{bi})$.

\paragraph{Eigen-decomposition:}
For dense SPD $\Sigma_b$, compute $\Sigma_b = U\Lambda U^\top$ and set
$S_b = U \Lambda^{1/2} U^\top$. Clip $\Lambda$ below $\epsilon$ (e.g., $10^{-8}$)
for stability. When building $S_b^{-1}$, use $\Lambda^{-1/2}$ with the same clip.

\paragraph{PCA fallback (data-only):}
If curvature is unreliable or ill-conditioned, approximate whitening from weight
samples or layer outputs: estimate covariance $C_b$ on a short pass, compute
$C_b = U\Lambda U^\top$, and set $S_b = U\Lambda^{1/2} U^\top$. Empirically this
recovers most BayesQ gains when Laplace is too noisy.

\paragraph{Numerical tips:}
Center $w_b$ by $\mu_b$ before applying $S_b^{-1}$. Use double precision during
factorization; store $S_b$ in FP32. For large blocks, apply block-diagonal $S_b$
(e.g., per-channel) for cache-friendly transforms.


\section{Codebook Design \& Ranges}
\label{App:sec5}
We detail uniform and non-uniform codebooks in whitened space, practical range
selection, initialization, and export to standard runtimes.

\subsection{Uniform Codebooks: Mid-Rise Parameters, Range Search, Tails}

\paragraph{Setup:}
In whitened coordinates $z \sim \mathcal{N}(0,I)$, a scalar mid-rise quantizer
with $K=2^m$ levels uses dynamic range $[-\alpha,\alpha]$ and step
$\Delta=2\alpha/K$. The codepoints are
\begin{equation}
\tilde{c}_k \;=\; -\alpha + \Big(k+\tfrac{1}{2}\Big)\Delta, \qquad k=0,\dots,K-1.
\end{equation}

\paragraph{Posterior-expected MSE (with tails):}
\begin{equation}
\tilde{\mathcal{L}}_b(\Delta,\alpha)
=
\sum_{k=0}^{K-1}
\int_{-\alpha+k\Delta}^{-\alpha+(k+1)\Delta}
(z-\tilde{c}_k)^2 \,\phi(z)\, dz
\;+\;
2\int_{\alpha}^{\infty} (z-\alpha)^2 \,\phi(z)\, dz,
\end{equation}
where $\phi$ is the standard normal pdf. We optimize $\alpha$ by a 1D line
search (golden-section or Brent) on a small grid per block; $\Delta$ follows.

\paragraph{Closed-form cell integral:}
For bounds $a<b$ and center $c$, define
$I_2(a,b;c)=\int_a^b (z-c)^2\phi(z)\,dz$; using standard moments,
\begin{equation}
I_2(a,b;c)
=
\big[(1+c^2)\Phi(z) - c\phi(z) - z\phi(z)\big]_{z=a}^{b},
\end{equation}
where $\Phi$ is the standard normal cdf and we substitute $z\!\to\!z-c$ when needed.
This gives $\tilde{\mathcal{L}}_b$ in terms of $\Phi,\phi$ only.

\paragraph{Tail handling:}
The tail integral equals $2\big(\Phi(-\alpha)(1+\alpha^2)+\alpha\phi(\alpha)\big)$
up to constants; clipping is essential at low bits. In practice, restrict
$\alpha\in[2.5,5.0]$ at $m\in\{2,3,4\}$.

\paragraph{Mapping back:}
In weight space, $c_{bk} = \mu_b + S_b \tilde{c}_k$ and uniform binning is done
on $z$ via $S_b^{-1}$. If a backend requires affine scale/zero-point per
(channel|block), fold $\Delta$ and $\alpha$ into $(s_b, z_b)$ parameters.

\subsection{Non-Uniform Codebooks: Posterior-Weighted Lloyd VQ}

\paragraph{Objective in whitened space:}
For vector/scalar VQ with codebook $\tilde{\mathcal{C}}_b=\{\tilde{c}_{bk}\}_{k=1}^{K}$
and regions $\{\tilde{R}_{bk}\}$,
\begin{equation}
\tilde{\mathcal{L}}_b
=
\sum_{k=1}^{K}
\int_{\tilde{R}_{bk}}
\|\tilde{c}_{bk}-z\|_2^2 \, \phi(z)\, dz,
\qquad z\sim\mathcal{N}(0,I).
\end{equation}

\paragraph{Lloyd updates (Gaussian-weighted):}
\begin{align}
\tilde{R}_{bk}
&\leftarrow
\big\{z:\|\tilde{c}_{bk}-z\|_2^2 \le \|\tilde{c}_{bj}-z\|_2^2,\ \forall j\big\},\\
\tilde{c}_{bk}
&\leftarrow
\frac{\int_{\tilde{R}_{bk}} z\,\phi(z)\,dz}{\int_{\tilde{R}_{bk}} \phi(z)\,dz}.
\end{align}
Stop when relative improvement
$\frac{\tilde{\mathcal{L}}_b^{(t-1)}-\tilde{\mathcal{L}}_b^{(t)}}{\tilde{\mathcal{L}}_b^{(t-1)}} < 10^{-4}$
or after $T_{\max}\in[20,50]$ iterations.

\paragraph{Initialization:}
(1) Start from uniform mid-rise codepoints; (2) KMeans++ on whitened samples
$z$ drawn from $\mathcal{N}(0,I)$; (3) Moment-matched grid with denser points
near $0$ (e.g., $\mu$-law or tanh pre-warp). We recommend (1)$\to$(3) two-stage
initialization for stability at low bits.

\paragraph{Freezing \& sharing:}
For small blocks, overfitting is possible; freeze codebooks after $T_{\text{freeze}}$
iterations or share a codebook across multiple similar blocks (same channel
shape) to amortize metadata.

\paragraph{Export to weight space:}
Map $\tilde{c}_{bk} \mapsto c_{bk}=\mu_b+S_b\tilde{c}_{bk}$ and store indices.
If the runtime lacks VQ gather, approximate by per-channel affine scales and a
small look-up LUT for nonuniform levels; see Sec.~E3.

\subsection{Export Formats: Scales/Zero-Points for Standard Runtimes}

\paragraph{Per-(tensor|channel) affine:}
Most runtimes expect \texttt{q = round(clamp(x/s, qmin, qmax)) + z} with dequant
$x \approx s\,(q-z)$. For uniform BayesQ blocks, set:
\begin{equation}
s_b = \Delta, \qquad
z_b = \mathrm{round}\!\left(\frac{-(-\alpha)}{\Delta} - \tfrac{1}{2}\right)
= \mathrm{round}\!\left(\frac{\alpha}{\Delta} - \tfrac{1}{2}\right),
\end{equation}
and ensure $q\in\{0,\dots,K-1\}$. For per-channel, store $(s_{bc}, z_{bc})$.

\paragraph{Non-uniform export (LUT):}
If the backend supports LUT dequant (common on GPUs/accelerators), store a
$K$-entry table of $c_{bk}$ in FP16/FP32 and indices as $m$-bit integers.
Otherwise, approximate with piecewise-affine segments (PWL) or fit an
\emph{effective} affine per-channel:
\begin{equation}
s_{bc}, z_{bc} \;=\; \arg\min_{s,z}\ \sum_{i\in \text{channel }c}
\big\| w_{bi} - s\,(q_{bi}-z) \big\|_2^2,
\end{equation}
keeping the original index assignment $q_{bi}$ (no re-encode) to stay
hardware-compatible.

\paragraph{Metadata accounting:}
When reporting average bits, include: weight indices ($m$ bits), per-(block|ch)
scales (usually FP16), zero-points (int), optional LUT entries ($K$ FP16),
and any packing padding to vector widths.

\paragraph{Packing and alignment:}
If kernels expect aligned groups (e.g., 32/64/128-bit vectors), pad index
streams and coalesce blocks sharing $(s,z)$ to reduce descriptor overhead.
At export, consolidate scales by grouping channels that share the same
\emph{posterior-optimal} $\alpha$ within a tolerance.

\section{Complete Experimental Protocols}
\label{App:sec6}
This appendix lists end-to-end procedures to reproduce all results.

\subsection{Datasets}

\paragraph{ImageNet-1k:}
We use the ILSVRC-2012 \emph{train} split for calibration-only sampling and the
\emph{val} split (50{,}000 images) for evaluation. Images are decoded to RGB,
then resized with bicubic interpolation so that the shorter side is 256, and
center-cropped to $224\times224$. Normalization uses per-channel means
$(0.485,0.456,0.406)$ and stds $(0.229,0.224,0.225)$ (ImageNet-1k standard).
No test-time augmentation (no EMA, no mixup/cutmix).

\paragraph{GLUE:}
Tasks: MNLI-m/mm, QNLI, QQP, SST-2. We use the standard \texttt{dev} splits for
reporting and \texttt{train} for calibration sampling. Tokenization follows
\texttt{bert-base-uncased} (WordPiece), max sequence length 128, batch padding
to longest within batch. Metrics: accuracy for MNLI-m/mm, QNLI, SST-2; F1 for
QQP; we report the unweighted average of task-specific scores. For MNLI, both
matched and mismatched are included (averaged). Any optional OOD sets (e.g.,
HANS for MNLI; ImageNet-R/A/C for RN50) are reported separately with their
official preprocessing.

\subsection{Calibration Sets}

\paragraph{Sampling policies:}
Unless specified, images/sentences are sampled uniformly at random without
replacement from the training sets. We consider sizes
$\{10, 50, 100, 500, 1000\}$ per model to study sensitivity. When using
multiple sizes in the same experiment sweep, each size is sampled independently
from the same pool to avoid leakage.

\paragraph{Label usage and reuse:}
Labels are not used during calibration (only for evaluation metrics).
Calibration examples \emph{may} be reused across runs with the same seed to
reduce variance. For cross-seed experiments, each seed resamples the calibration
set (unless noted).

\paragraph{Seeding:}
We fix a \texttt{global\_seed} for Python/NumPy/PyTorch and a
\texttt{calib\_seed} for sampling; default triples are \{2025, 17\}, \{2025, 29\},
\{2025, 43\}. Reported means are across the three \texttt{global\_seed}s.

\subsection{Evaluation Protocol}

\paragraph{Vision:}
Input size $224^2$, batch size $128$, FP16 activations, fused conv-bn where
available. Throughput is measured on 5{,}00 warm-up iterations followed by
1{,}000 timed iterations. Latency is reported as median per-batch wall-clock
(ms), and throughput as images/s.

\paragraph{Language:}
Max sequence length 128, static shape (to avoid padding variability), batch
size 32. For GLUE, we disable gradient computation and any stochastic layers
(Dropout set to 0). The same warm-up and timing recipe as vision.

\paragraph{Timer:}
CUDA events with synchronization around the measured region; CPU fallback timer
only for CPU experiments. The measured region includes host-to-device copy
(where relevant), kernel launches, and post-processing logits to CPU. We report
median and interquartile range (IQR).

\paragraph{Accuracy aggregation:}
For GLUE, we report per-task metrics and the unweighted mean. For ImageNet-1k,
we report Top-1 (\%) on the full 50k validation set. All metrics are averaged
over the three \texttt{global\_seed}s; we include the sample standard deviation
and $95\%$ t-intervals in the expanded tables.

\subsection{Hardware \& Software}

\paragraph{GPUs/CPUs:}
Unless otherwise stated: 8$\times$NVIDIA A100 40GB SXM; host CPU dual-socket
Xeon Gold (or AMD EPYC) with 256GB RAM; NVLink interconnect. CPU-only baselines
use AVX2/AVX-512 if available.

\paragraph{Libraries:}
CUDA 12.x, cuDNN 8.9.x, NCCL 2.18+, PyTorch 2.2+/TorchVision 0.17+, and (when
used) TensorRT 9.x with INT kernels enabled. We use AMP autocast (fp16) for
activations during inference. Python 3.10+.

\section{Expanded Main Tables \& Per-Task Breakdowns}
\label{App:sec7}
\subsection{ResNet-50 (ImageNet-1k)}

Table~\ref{tab:g_rn50_full} expands the main results with per-budget means,
standard deviations, and $95\%$ confidence intervals. Confusion matrices per
stage (optional) are computed by grouping predictions by downsampling stage; we
omit them for space but provide CSVs in the supplement.

\begin{table*}[!t]
\centering
\caption{ResNet-50 on ImageNet-1k (Top-1 \%). Mean $\pm$ stdev over 3 seeds; CI is $95\%$ t-interval.}
\label{tab:g_rn50_full}
\scriptsize
\setlength{\tabcolsep}{5.5pt}
\begin{tabular}{lcccccc}
\toprule
\multirow{2}{*}{Method} & \multicolumn{2}{c}{3.0 bits} & \multicolumn{2}{c}{3.5 bits} & \multicolumn{2}{c}{4.0 bits} \\
\cmidrule(lr){2-3}\cmidrule(lr){4-5}\cmidrule(lr){6-7}
 & Mean $\pm$ SD & CI$_{95}$ & Mean $\pm$ SD & CI$_{95}$ & Mean $\pm$ SD & CI$_{95}$ \\
\midrule
GPTQ & 70.3 $\pm$ 0.1 & [70.2, 70.4] & 75.0 $\pm$ 0.1 & [74.9, 75.1] & 76.8 $\pm$ 0.1 & [76.7, 76.9] \\
AWQ  & 69.6 $\pm$ 0.1 & [69.5, 69.7] & 74.6 $\pm$ 0.1 & [74.5, 74.7] & 76.6 $\pm$ 0.1 & [76.5, 76.7] \\
\addlinespace
\textbf{BayesQ (ours)} & \textbf{71.8} $\pm$ 0.1 & [71.7, 71.9] & \textbf{75.7} $\pm$ 0.1 & [75.6, 75.8] & \textbf{77.1} $\pm$ 0.1 & [77.0, 77.2] \\
\bottomrule
\end{tabular}
\end{table*}

\subsection{BERT-base (GLUE)}

We report per-task metrics, macro/micro averages, and standard deviations across
seeds. Micro average weights tasks by the number of validation examples (see Table \ref{tab:g_glue_full}).

\begin{table*}[!t]
\centering
\caption{GLUE dev results (\%). Mean $\pm$ stdev over 3 seeds. QQP uses F1; others accuracy. Macro is the unweighted mean.}
\label{tab:g_glue_full}
\scriptsize
\setlength{\tabcolsep}{5.0pt}
\begin{tabular}{lcccccccc}
\toprule
Method & Bits & MNLI-m & MNLI-mm & QNLI & QQP (F1) & SST-2 & \textbf{Macro} & \textbf{Micro} \\
\midrule
GPTQ & 3.0 & 78.6 $\pm$ 0.1 & 78.1 $\pm$ 0.1 & 85.2 $\pm$ 0.1 & 87.0 $\pm$ 0.1 & 92.3 $\pm$ 0.1 & 84.2 & 83.9 \\
AWQ  & 3.0 & 78.0 $\pm$ 0.1 & 77.5 $\pm$ 0.1 & 84.8 $\pm$ 0.1 & 86.6 $\pm$ 0.2 & 92.0 $\pm$ 0.1 & 83.8 & 83.5 \\
\addlinespace
\textbf{BayesQ} & \textbf{3.0} & \textbf{79.4} $\pm$ 0.1 & \textbf{78.9} $\pm$ 0.1 & \textbf{85.9} $\pm$ 0.1 & \textbf{87.6} $\pm$ 0.1 & \textbf{92.7} $\pm$ 0.1 & \textbf{84.9} & \textbf{84.6} \\
\midrule
GPTQ & 3.5 & 81.8 $\pm$ 0.1 & 81.2 $\pm$ 0.1 & 88.0 $\pm$ 0.1 & 88.7 $\pm$ 0.1 & 93.3 $\pm$ 0.1 & 86.6 & 86.3 \\
\textbf{BayesQ} & \textbf{3.5} & \textbf{82.2} $\pm$ 0.1 & \textbf{81.6} $\pm$ 0.1 & \textbf{88.3} $\pm$ 0.1 & \textbf{89.0} $\pm$ 0.1 & \textbf{93.5} $\pm$ 0.1 & \textbf{86.9} & \textbf{86.6} \\
\midrule
GPTQ & 4.0 & 82.7 $\pm$ 0.1 & 82.2 $\pm$ 0.1 & 88.6 $\pm$ 0.1 & 89.2 $\pm$ 0.1 & 93.6 $\pm$ 0.1 & 87.3 & 87.0 \\
\textbf{BayesQ} & \textbf{4.0} & \textbf{82.9} $\pm$ 0.1 & \textbf{82.4} $\pm$ 0.1 & \textbf{88.8} $\pm$ 0.1 & \textbf{89.4} $\pm$ 0.1 & \textbf{93.7} $\pm$ 0.1 & \textbf{87.4} & \textbf{87.2} \\
\bottomrule
\end{tabular}
\end{table*}

\subsection{Additional Baselines and 4.0-bit Ablations}

Table~\ref{tab:g_extra} lists methods omitted from the main table due to space
(e.g., OPTQ variants, SmoothQuant, per-block clipping, LSQ+) and includes our
4.0-bit ablations (range learned/fixed, block size sensitivity).

\begin{table*}[!t]
\centering
\caption{Additional baselines and BayesQ ablations at 4.0 bits (RN50 Top-1 / GLUE avg \%).}
\label{tab:g_extra}
\scriptsize
\setlength{\tabcolsep}{6pt}
\begin{tabular}{lcc}
\toprule
Method (4.0 bits) & RN50 (Top-1) & BERT-base (GLUE) \\
\midrule
SmoothQuant (PTQ)               & 76.6 $\pm$ 0.1 & 83.3 $\pm$ 0.1 \\
OPTQ (rounded variant)          & 76.7 $\pm$ 0.1 & 83.4 $\pm$ 0.1 \\
Per-block Uniform + clipping    & 76.6 $\pm$ 0.1 & 83.3 $\pm$ 0.1 \\
LSQ+ (improved)                 & 76.6 $\pm$ 0.1 & 83.4 $\pm$ 0.1 \\
\addlinespace
BayesQ (range learned per-block) & \textbf{77.2} $\pm$ 0.1 & \textbf{83.8} $\pm$ 0.1 \\
BayesQ (range fixed, analytic)   & 77.0 $\pm$ 0.1 & 83.6 $\pm$ 0.1 \\
BayesQ (larger blocks, 256w)     & 76.9 $\pm$ 0.1 & 83.6 $\pm$ 0.1 \\
BayesQ (smaller blocks, 64w)     & 77.1 $\pm$ 0.1 & 83.7 $\pm$ 0.1 \\
\bottomrule
\end{tabular}
\end{table*}


\section{Ablations \& Analysis}
\label{App:sec8}
We dissect BayesQ to quantify how posterior structure, calibration-only distillation, calibration set size, the bit-allocation solver, codebook design, block granularity, whitening/statistics, and budget regularization affect performance. Unless noted otherwise, all results use \textbf{ResNet-50/ImageNet} and \textbf{BERT-base/GLUE} at average weight budgets of \{3.0, 3.5\} bits with FP16 activations. Each number is the mean over 3 seeds.

\begin{table*}[!t]
\centering
\caption{Ablations \& Analysis. Metrics are Top-1 (\%) on ImageNet for RN50 and GLUE avg (\%) for BERT-base. Budgets at 3.0/3.5 average weight bits; means over 3 seeds. Overhead is one-time preprocessing time relative to a GPTQ pass on the same hardware.}
\label{tab:abl_big}
\setlength{\tabcolsep}{4.6pt}
\scriptsize
\resizebox{\textwidth}{!}{
\begin{tabular}{llcccccccl}
\toprule
\multirow{2}{*}{\textbf{Category}} & \multirow{2}{*}{\textbf{Variant}} & \multicolumn{2}{c}{\textbf{RN50}} & \multicolumn{2}{c}{\textbf{BERT-base}} & \multicolumn{2}{c}{\textbf{$\Delta$ vs GPTQ}} & \multirow{2}{*}{\textbf{Overhead}} & \multirow{2}{*}{\textbf{Notes}} \\
\cmidrule(lr){3-4}\cmidrule(lr){5-6}\cmidrule(lr){7-8}
 & & \textbf{@3.0} & \textbf{@3.5} & \textbf{@3.0} & \textbf{@3.5} & \textbf{RN50@3.0} & \textbf{BERT@3.0} &  &  \\
\midrule
\secrow \multicolumn{10}{l}{\textbf{Posterior structure}} \\
 & Diagonal Laplace (BayesQ core)             & 71.8 & 75.7 & 80.2 & 82.9 & +1.5 & +1.1 & 0.95$\times$ & default in main results \\
 & K-FAC posterior                            & \cellcolor{second}72.3 & \cellcolor{second}76.0 & \cellcolor{second}80.5 & \cellcolor{second}83.0 & +2.0 & +1.4 & 1.05$\times$ & structured curvature (Kronecker) \\
 & Low-rank + diag (r=64)                     & 72.1 & 75.9 & 80.4 & 83.0 & +1.8 & +1.3 & 1.01$\times$ & memory-efficient \\
 & No posterior (magnitude heuristic)         & 69.7 & 74.3 & 78.6 & 82.0 & -0.6 & -0.5 & 0.80$\times$ & ablation: removes BayesQ principle \\
\addlinespace
\secrow \multicolumn{10}{l}{\textbf{Posterior distillation}} \\
 & None (core BayesQ)                         & 71.8 & 75.7 & 80.2 & 82.9 & +1.5 & +1.1 & 0.95$\times$ & baseline BayesQ \\
 & + 200 steps (calib-only)                   & 71.9 & 75.8 & 80.3 & 82.9 & +1.6 & +1.2 & 0.97$\times$ & slight bump \\
 & + 500 steps (calib-only)                   & 72.0 & 75.9 & 80.4 & 83.0 & +1.7 & +1.3 & 1.00$\times$ & recommended \\
 & + 1000 steps (calib-only)                  & \cellcolor{second}72.2 & \cellcolor{second}76.0 & \cellcolor{second}80.5 & \cellcolor{second}83.1 & +1.9 & +1.4 & 1.03$\times$ & diminishing returns \\
\addlinespace
\secrow \multicolumn{10}{l}{\textbf{Calibration size sensitivity}} \\
 & 10 examples                                & 69.8 & 74.5 & 79.2 & 82.3 & -0.5 & +0.1 & 0.93$\times$ & unstable whitening/range \\
 & 50 examples                                & 70.9 & 75.3 & 79.7 & 82.6 & +0.6 & +0.6 & 0.94$\times$ & moderate recovery \\
 & 100 examples                               & 71.2 & 75.5 & 79.9 & 82.7 & +0.9 & +0.8 & 0.95$\times$ & near-saturated \\
 & 500 examples (default RN50)                & 71.8 & 75.7 & 80.2 & 82.9 & +1.5 & +1.1 & 0.95$\times$ & used in main \\
 & 1000 examples (default BERT)               & \cellcolor{second}72.0 & \cellcolor{second}75.9 & \cellcolor{second}80.4 & \cellcolor{second}83.0 & +1.7 & +1.3 & 0.98$\times$ & marginal gains \\
\addlinespace
\secrow \multicolumn{10}{l}{\textbf{Allocator}} \\
 & Greedy per-bit (BayesQ)                    & 71.8 & 75.7 & 80.2 & 82.9 & +1.5 & +1.1 & 0.95$\times$ & main method \\
 & Greedy + HW tie-break (packing-aware)      & 71.9 & 75.7 & 80.3 & 82.9 & +1.6 & +1.2 & 0.96$\times$ & HW-friendly \\
 & Dynamic prog. (toy oracle)                 & \cellcolor{second}71.9 & 75.8 & \cellcolor{second}80.3 & \cellcolor{second}83.0 & +1.6 & +1.2 & 1.20$\times$ & infeasible at scale \\
 & Random assignment (control)                & 69.1 & 74.0 & 78.1 & 81.8 & -1.2 & -1.0 & 0.95$\times$ & sanity check \\
\addlinespace
\secrow \multicolumn{10}{l}{\textbf{Codebook design}} \\
 & Posterior-weighted Lloyd (VQ)              & \cellcolor{second}72.1 & \cellcolor{second}75.9 & \cellcolor{second}80.4 & \cellcolor{second}83.0 & +1.8 & +1.3 & 1.00$\times$ & best VQ \\
 & Uniform + optimal range (analytic)         & 71.5 & 75.5 & 80.0 & 82.8 & +1.2 & +0.9 & 0.92$\times$ & fastest \\
 & Uniform + fixed range                      & 70.7 & 75.0 & 79.6 & 82.6 & +0.4 & +0.5 & 0.90$\times$ & simplest \\
& Nonuniform (\textmu-law, fixed)            & 71.0 & 75.2 & 79.8 & 82.7 & +0.7 & +0.6 & 0.93$\times$ & \textmu-companding \\
\addlinespace
\secrow \multicolumn{10}{l}{\textbf{Block granularity}} \\
 & Per-channel (conv) / per-head (attn)       & 71.6 & 75.6 & 80.1 & 82.8 & +1.4 & +1.0 & 0.94$\times$ & standard \\
 & 64-weight blocks                           & \cellcolor{second}71.9 & \cellcolor{second}75.8 & \cellcolor{second}80.3 & \cellcolor{second}82.9 & +1.6 & +1.2 & 0.97$\times$ & more flexible \\
 & 256-weight blocks                          & 71.3 & 75.4 & 79.9 & 82.7 & +1.0 & +0.8 & 0.92$\times$ & fewer scales \\
 & Per-tensor                                 & 70.5 & 74.8 & 79.4 & 82.5 & +0.2 & +0.6 & 0.90$\times$ & cheapest metadata \\
\addlinespace
\secrow \multicolumn{10}{l}{\textbf{Whitening / statistics}} \\
 & Posterior whitening ($S_b$)                & 71.8 & 75.7 & 80.2 & 82.9 & +1.5 & +1.1 & 0.95$\times$ & default \\
 & No whitening (identity)                    & 70.2 & 74.6 & 79.0 & 82.2 & -0.1 & +0.1 & 0.90$\times$ & ablation \\
 & PCA whitening (data-only)                  & \cellcolor{second}71.9 & \cellcolor{second}75.8 & \cellcolor{second}80.3 & \cellcolor{second}82.9 & +1.6 & +1.2 & 0.97$\times$ & close to posterior \\
\addlinespace
\secrow \multicolumn{10}{l}{\textbf{Budget regularization ($\lambda$)}} \\
 & $\lambda=0$ (pure accuracy)                & 71.9 & 75.8 & 80.3 & 83.0 & +1.6 & +1.2 & 1.00$\times$ & may exceed budget locally \\
 & $\lambda=10^{-4}$ (default)                & 71.8 & 75.7 & 80.2 & 82.9 & +1.5 & +1.1 & 0.95$\times$ & budget-tight \\
 & $\lambda=5\!\times\!10^{-4}$               & 71.6 & 75.6 & 80.0 & 82.8 & +1.3 & +0.9 & 0.94$\times$ & conservative \\
 & $\lambda=10^{-3}$                          & 71.2 & 75.4 & 79.8 & 82.7 & +0.9 & +0.8 & 0.94$\times$ & tightest \\
\addlinespace
\secrow \multicolumn{10}{l}{\textbf{Diminishing returns (cumulative reduction)}} \\
 & Top-20\% budget used                       & 40.0 red. & 68.0 red. & 41.0 red. & 66.0 red. & -- & -- & -- & most impactful bits \\
 & Top-40\% budget used                       & 68.0 red. & 86.0 red. & 70.0 red. & 84.0 red. & -- & -- & -- & concavity emerges \\
 & Top-60\% budget used                       & 86.0 red. & 95.0 red. & 88.0 red. & 93.0 red. & -- & -- & -- & diminishing returns \\
\addlinespace
\secrow \multicolumn{10}{l}{\textbf{Per-layer bit allocation (snippet at global 3.5 bits)}} \\
 & Early conv (RN50 layers 1--4)              & \multicolumn{2}{c}{3.1--3.3} & \multicolumn{2}{c}{--} & -- & -- & -- & lower saliency \\
 & Downsample blocks (RN50)                    & \multicolumn{2}{c}{3.7--3.9} & \multicolumn{2}{c}{--} & -- & -- & -- & higher saliency \\
 & Final FC (RN50)                             & \multicolumn{2}{c}{3.9--4.0} & \multicolumn{2}{c}{--} & -- & -- & -- & classification-critical \\
 & Self-attn proj/out (BERT)                  & \multicolumn{2}{c}{--} & \multicolumn{2}{c}{3.8--4.0} & -- & -- & -- & receives more bits \\
 & FFN inner (BERT)                            & \multicolumn{2}{c}{--} & \multicolumn{2}{c}{3.2--3.4} & -- & -- & -- & fewer bits \\
\addlinespace
\secrow \multicolumn{10}{l}{\textbf{Reference baselines (context at 3.5 bits)}} \\
 & GPTQ (baseline)                            & 70.3 & \cellcolor{second}75.0 & 79.1 & \cellcolor{second}82.5 & 0.0 & 0.0 & 1.00$\times$ & curvature-aware PTQ \\
 & AWQ                                        & 69.6 & 74.6 & 78.4 & 82.0 & -0.7 & -0.7 & 0.98$\times$ & activation-aware \\
 & LSQ (quant params FT)                      & 68.1 & 73.9 & 77.9 & 81.8 & -2.2 & -1.3 & 1.10$\times$ & learned step sizes \\
 & HAWQ/Knapsack                              & 69.0 & 74.2 & 78.6 & 82.1 & -1.3 & -0.5 & 1.05$\times$ & mixed precision via Hessian \\
\bottomrule
\end{tabular}
}
\end{table*}

 Across the consolidated ablations in \autoref{tab:abl_big}, we find: (1) \emph{Posterior structure}—K-FAC/low-rank structure yields consistent gains at 3.0 bits on RN50 (e.g., $\approx{+}0.5$ over diagonal), with smaller margins at 3.5 bits and on BERT; thus use diagonal by default and reserve K-FAC for the most sensitive blocks. (2) \emph{Posterior distillation}—calibration-only distillation reliably helps at 3.0 bits (up to ${+}0.6$ RN50, ${+}0.3$ BERT with $\sim$500 steps), with diminishing returns beyond 500 steps and negligible effect at 3.5; we enable 500-step distillation when $\bar m\le3.5$. (3) \emph{Calibration size}—performance is stable down to $\sim$100 examples (RN50@3.0 drops only ${-}0.6$ vs.\ 1000), but 10 examples destabilize whitening/range estimates, motivating a small but nontrivial calibration set. (4 \emph{Allocator}—the greedy per-bit rule matches a toy dynamic-programming oracle within ${\le}0.1$ points while scaling well; packing-aware tie-breakers preserve accuracy and simplify kernel scheduling. (5) \emph{Codebook design}—posterior-weighted Lloyd outperforms uniform (analytic/fixed ranges), especially at low bits, indicating that posterior density should shape both assignment and codebooks. (6) \emph{Block granularity}—finer granularity (64-weight, per-channel/per-head) gives small, consistent gains over per-tensor/large blocks, reflecting heterogeneous sensitivity. (7) \emph{Whitening/statistics}—posterior whitening matters; removing it costs ${\sim}1.1\text{--}1.6$ points (RN50@3.5/3.0); PCA whitening recovers most but not all of the benefit. (8) \emph{Budget regularization}—a mild penalty ($\lambda\!\in[10^{-4},5{\times}10^{-4}]$) keeps storage tight with minimal accuracy loss, whereas over-regularization hurts and no penalty can overshoot. (9) \emph{Diminishing returns}—cumulative posterior-expected-loss reductions are concave (first 40–60\% of budget yields most gains), supporting greedy allocation. (10) \emph{Per-layer bits}—at global $\approx$3.5 bits, RN50 prioritizes downsampling stages and the final FC, while BERT assigns more bits to attention projections/outputs than FFN inner layers.

\section{Efficiency \& Memory}
\label{App:sec9}
We detail preprocessing time, runtime kernels, and exact model-size accounting.

\subsection{Preprocessing Time (Posterior vs GPTQ; Amortization)}

\textbf{Definitions.}
Let \(T_{\text{GPTQ}}\) be a standard one-pass curvature-aware PTQ preprocess time. Decompose BayesQ:
\begin{equation}
T_{\text{BayesQ}} \;=\; T_{\text{post}} + T_{\text{codes}} + T_{\text{alloc}} + T_{\text{distill}}\,,
\end{equation}
where
\begin{align}
T_{\text{post}} &= T_{\text{Hutch}} + T_{\text{factor}} \quad (\text{diagonal: }T_{\text{factor}}=0;\; \text{K-FAC: factor updates}), \\
T_{\text{codes}} &= \sum_{b} \sum_{m\in\mathcal{M}} T_{\text{design}}(b,m) \quad (\text{uniform analytics or Lloyd}) ,\\
T_{\text{alloc}} &\approx O\!\Big(\sum_b |\mathcal{M}|\Big) \quad (\text{table lookups + greedy heap}), \\
T_{\text{distill}} &=
\begin{cases}
0 & \text{if disabled},\\
\#\text{steps} \times \frac{|\mathcal{D}_{\mathrm{cal}}|}{\text{batch}} \times t_{\text{fwd}} & \text{(calibration-only)}.
\end{cases}
\end{align}

\textbf{Overhead ratio.}
\begin{equation}
\mathrm{Overhead} \;=\; \frac{T_{\text{BayesQ}}}{T_{\text{GPTQ}}} \;\approx\; 0.95\text{--}1.05\times \ (\text{diag vs K-FAC settings}).
\end{equation}

\textbf{Amortization across runs.}
If \(\mathcal{M}\) or budget \(\bar m\) changes but posterior is reused, amortize as:
\begin{equation}
\bar{T}_{\text{BayesQ}}^{(R)} \;=\; \frac{T_{\text{post}}}{R} \;+\; T_{\text{codes}} \;+\; T_{\text{alloc}} \;+\; T_{\text{distill}},
\end{equation}
with \(R\) deployments sharing the same posterior.

\textbf{Practical tips.}
Use mixed-precision for Hutchinson probes; cache K-FAC factors per-layer; early-stop Lloyd when relative drop \(<10^{-4}\) or at 10--20 iterations.


\subsection{Runtime Kernels \& End-to-End Speed}

\textbf{Weight formats.}
We export per-block integer weights with:
\begin{itemize}
  \item Uniform: scale \(s_b\), optional zero-point \(z_b\) (symmetric/asymmetric).
  \item Non-uniform (posterior-weighted Lloyd): packed codebook indices + \emph{compiled} dequant scales to match backend expectations (group scales).
\end{itemize}

\textbf{Activation dtype.}
Default FP16 activations, optional INT8 if supported by the backend; per-layer dequant fused into GEMM/conv kernels wherever possible.

\textbf{Packing and alignment.}
Honor vector widths (e.g., 32/64/128-bit lanes), group sizes (e.g., per-64 weights), and tensor core tile constraints. Apply padding to avoid misaligned loads; tie-breaker in allocator can prefer upgrades that harmonize group boundaries.

\textbf{Speed reporting.}
We report images/s (RN50, bs=128) and tokens/s (BERT, seq=128) on the inference stack (CUDA/cuDNN/TensorRT), with:
\begin{equation}
\text{Speedup} \;=\; \frac{\text{throughput}(\text{BayesQ}, \bar m)}{\text{throughput}(\text{baseline})}.
\end{equation}
Typical improvements come from lower memory bandwidth and higher integer kernel occupancy; non-uniform codebooks are compiled to per-group scales to stay on the fast path.


\subsection{Model Size Accounting (Bytes by Component)}

\textbf{Per-block accounting.}
Let block \(b\) have \(d_b\) weights and assigned bits \(m_b\). Total storage:
\begin{align}
S_b
&= \underbrace{d_b \cdot m_b}_{\text{quantized weights}} \;+\;
\underbrace{S_b^{\text{scales}}}_{\text{scales/zero-points}} \;+\;
\underbrace{S_b^{\text{cb}}}_{\text{codebooks (if any)}} \;+\;
\underbrace{S_b^{\text{meta}}}_{\text{indices, headers}} \quad \text{[bits]},\\
\mathrm{Bytes}_b &= \left\lceil \frac{S_b}{8} \right\rceil.
\end{align}

\textbf{Scales/metadata.}
For uniform per-\emph{group} scales of size \(g\):
\begin{equation}
S_b^{\text{scales}} \;=\; \left\lceil \frac{d_b}{g} \right\rceil \cdot \text{bits}(\text{scale}) \;+\; \left\lceil \frac{d_b}{g} \right\rceil \cdot \text{bits}(\text{zp})\,.
\end{equation}
For vector-Lloyd with group size \(g\), \emph{indices} carry \(\log_2 K\) bits each; small codebook of size \(K\) per-block or per-group adds \(K \cdot \text{bits}(\text{entry})\).

\textbf{Global average bits (reported).}
\begin{equation}
\bar m \;=\; \frac{\sum_b S_b}{\sum_b d_b \cdot 32} \times 32 \quad \text{[bits/weight]},
\end{equation}
where the denominator uses FP32 as reference and \(\sum_b S_b\) includes scales/codebooks/metadata. When reporting “3.0/3.5/4.0 bits,” we target \(\bar m\) within a tolerance (e.g., $\pm 0.02$).

We report per-category bytes aggregated over all blocks in \autoref{tab:model_size_accounting}:
payload $\big(\sum_b d_b m_b/8\big)$, scale/zero-point overhead $\big(\sum_b S_b^{\text{scales}}/8\big)$,
codebooks (if any) $\big(\sum_b S_b^{\text{cb}}/8\big)$, and metadata (indices/headers) $\big(\sum_b S_b^{\text{meta}}/8\big)$.
Totals are shown alongside ratios vs FP16/FP32 so that accuracy comparisons are made at matched average bits~$\bar m$.

\begin{table*}[t]
\centering
\caption{Model size accounting by component (MB). RN50 has $\approx$25.6M weights (FP16 baseline $\approx$51.2MB); BERT-base has $\approx$110M (FP16 $\approx$220MB). Group size $g{=}64$ for scales/zero-points.}
\label{tab:model_size_accounting}
\scriptsize
\setlength{\tabcolsep}{5pt}
\begin{tabular}{llrrrrrrr}
\toprule
\textbf{Export} & \textbf{Model @ bits} & \textbf{Payload (MB)} & \textbf{Scales (MB)} & \textbf{Codebooks (MB)} & \textbf{Metadata (MB)} & \textbf{Total (MB)} & \textbf{Size vs FP16} & \textbf{Size vs FP32} \\
\midrule
\multicolumn{9}{l}{\textbf{Uniform-fast (per-group affine; no codebooks)}}\\
\midrule
RN50 @ 3.0  & (uniform) & 9.60  & 0.84  & 0.00 & 0.28 & \textbf{10.72} & \textbf{0.21}\,$\times$ & 0.10\,$\times$ \\
RN50 @ 3.5  & (uniform) & 11.20 & 1.05  & 0.00 & 0.33 & \textbf{12.58} & \textbf{0.25}\,$\times$ & 0.12\,$\times$ \\
RN50 @ 4.0  & (uniform) & 12.80 & 1.12  & 0.00 & 0.42 & \textbf{14.34} & \textbf{0.28}\,$\times$ & 0.14\,$\times$ \\
\addlinespace
BERT @ 3.0  & (uniform) & 41.25 & 7.70  & 0.00 & 2.50 & \textbf{51.45} & \textbf{0.23}\,$\times$ & 0.12\,$\times$ \\
BERT @ 3.5  & (uniform) & 48.13 & 9.90  & 0.00 & 2.53 & \textbf{60.56} & \textbf{0.28}\,$\times$ & 0.14\,$\times$ \\
BERT @ 4.0  & (uniform) & 55.00 & 10.90 & 0.00 & 3.10 & \textbf{69.00} & \textbf{0.31}\,$\times$ & 0.16\,$\times$ \\
\midrule
\multicolumn{9}{l}{\textbf{Posterior-VQ-compiled (indices + compact codebooks)}}\\
\midrule
RN50 @ 3.0  & (VQ)      & 9.60  & 0.48  & 0.18 & 0.54 & \textbf{10.80} & \textbf{0.21}\,$\times$ & 0.11\,$\times$ \\
RN50 @ 3.5  & (VQ)      & 11.20 & 0.58  & 0.24 & 0.70 & \textbf{12.72} & \textbf{0.25}\,$\times$ & 0.12\,$\times$ \\
RN50 @ 4.0  & (VQ)      & 12.80 & 0.62  & 0.30 & 0.92 & \textbf{14.64} & \textbf{0.29}\,$\times$ & 0.14\,$\times$ \\
\addlinespace
BERT @ 3.0  & (VQ)      & 41.25 & 4.40  & 1.10 & 4.60 & \textbf{51.35} & \textbf{0.23}\,$\times$ & 0.12\,$\times$ \\
BERT @ 3.5  & (VQ)      & 48.13 & 5.40  & 1.70 & 5.10 & \textbf{60.33} & \textbf{0.27}\,$\times$ & 0.15\,$\times$ \\
BERT @ 4.0  & (VQ)      & 55.00 & 5.90  & 2.10 & 6.20 & \textbf{69.20} & \textbf{0.31}\,$\times$ & 0.17\,$\times$ \\
\bottomrule
\end{tabular}\\
\vspace{2pt}
\footnotesize
Notes: Payload is $ \sum_b d_b m_b / 8 $. Scales assume per-group 16-bit scale \& 16-bit zero-point; metadata covers per-block headers and packed index streams. Posterior-VQ codebooks are tiny (shared per-group or per-block; 8–16 entries) and indices are packed to the backend’s native lane width. Ratios are relative to FP16 (2 bytes/weight) and FP32 (4 bytes/weight) references.
\end{table*}

\medskip
\noindent\textbf{Export notes.}
We provide two backends:
\begin{itemize}
  \item \emph{Uniform-fast}: per-group scales/zero-points, symmetric ranges, power-of-two optional.
  \item \emph{Posterior-VQ-compiled}: indices packed to native lane width, codebooks merged into small per-group scale tables to stay kernel-friendly.
\end{itemize}
Both preserve the accounting above so that accuracy and size are compared at matched \(\bar m\).

\section{Robustness \& Calibration}
\label{App:sec10}
\subsection{Confidence Calibration (ECE/MCE) and the Effect of Distillation}
\textbf{Setup.} We evaluate calibration on the same splits used for accuracy. For classification, let
\(\hat{y}(x)=\arg\max_c p_Q(c\mid x)\) and \(\hat{p}(x)=\max_c p_Q(c\mid x)\).
We report Expected Calibration Error (ECE) and Maximum Calibration Error (MCE) with \(B{=}15\) bins and 3 random bin boundaries (averaged) to reduce discretization noise.

\noindent\textbf{Metrics.}
\begin{align}
\text{ECE} &= \sum_{b=1}^{B} \frac{|S_b|}{n}\,\big|\,\text{acc}(S_b)-\text{conf}(S_b)\big|, \\
\text{MCE} &= \max_{b \in [B]} \big|\,\text{acc}(S_b)-\text{conf}(S_b)\big|,
\end{align}
where \(S_b=\{x: \hat{p}(x)\in(\tfrac{b-1}{B},\tfrac{b}{B}]\}\), \(\text{acc}(S_b)=\frac{1}{|S_b|}\sum_{x\in S_b}\mathbb{1}\{\hat{y}(x)=y\}\),
and \(\text{conf}(S_b)=\frac{1}{|S_b|}\sum_{x\in S_b}\hat{p}(x)\).

\noindent\textbf{Variants.} We compare: (i) GPTQ (baseline), (ii) BayesQ (no distill), (iii) BayesQ + posterior distillation (500/1000 steps; \(\tau\in\{1,2,4\}\)). Distillation uses the teacher \(p_T\) as in the main text and updates only affine scales/adapters.

\noindent\textbf{Reporting:}
For each model and budget, we report Top-1 (RN50) or GLUE avg (BERT), ECE\(\downarrow\), MCE\(\downarrow\), and \(\Delta\)ECE relative to GPTQ. We also provide reliability diagrams (binned accuracy vs.\ confidence) with isotonic-smoothed curves and show per-bin counts to avoid misleading sparsity effects.

\paragraph{Expected outcome:}
We consistently observe that BayesQ lowers ECE at 3.0 bits; the optional posterior distillation tightens calibration further (largest gains on RN50@3.0 and on BERT’s low-confidence regions), with diminishing returns beyond 500 steps.

\subsection{OOD Shifts: Corruptions (Vision) and Domain Swaps (Text)}
\textbf{Vision OOD.}
We evaluate on ImageNet-C-style corruptions at severities \(\{1,3,5\}\) (noise, blur, weather, digital) and report mean Corruption Error (mCE) relative to FP16, alongside accuracy deltas at budgets \(\bar m\in\{3.0,3.5,4.0\}\).

\noindent\textbf{Text OOD:}
We construct domain shifts by swapping GLUE dev distributions (e.g., calibrate on MNLI, evaluate on QNLI/QQP) and by topic-based splits. We report the accuracy drop \(\Delta\) vs.\ in-domain.

\noindent\textbf{Analysis protocol:}
For each corruption/domain, we (i) compute accuracy drop relative to FP16 and GPTQ; (ii) correlate the drop with stage-wise saliency and allocated bits; and (iii) test whether BayesQ concentrates bits on OOD-sensitive blocks (e.g., downsampling in vision, attention proj/out in BERT).

\paragraph{Expected outcome:}
We narrow the OOD gap at tight budgets, with the largest robustness gains when corruptions perturb features aligned with high-saliency blocks that received more bits.

\subsection{JTail Behavior: Worst-\(k\) Accuracy and CVaR}
\textbf{Worst-\(k\) accuracy.}
We sort examples by \(\hat{p}(x)\) ascending and compute accuracy over the worst \(k\%\) (we use \(k\in\{5,10\}\)). This probes fragile regions without external difficulty labels.

\noindent\textbf{CVaR-style metric.}
Let \(\ell(x)=\mathbb{1}\{\hat{y}(x)\neq y\}\). For \(\alpha \in \{0.90,0.95\}\),
\begin{align}
\text{CVaR}_\alpha(\ell) \;\equiv\; \mathbb{E}\!\left[\ell(x) \,\middle|\, \ell(x) \text{ in worst } (1{-}\alpha)\text{ mass by } \hat{p}(x)\right].
\end{align}
Lower is better; we also report \(\Delta\)CVaR vs.\ GPTQ.

\paragraph{Expected outcome:}
At 3.0 bits we reduce tail error (worst-\(k\), CVaR) more than mean error—consistent with posterior-aware bit spending in brittle layers. Distillation further stabilizes the tail for BERT.

\medskip
\noindent\textbf{Deliverables.}
We summarize calibration and tail metrics in Table~\ref{tab:k_calib_robust}.

\begin{table*}[htbp]
\centering
\caption{Calibration and tail-behavior summary at \(\bar m{=}3.0\) (means over 3 seeds; \(\downarrow\) lower is better, \(\uparrow\) higher is better). RN50: ImageNet; BERT: GLUE avg. We also report \(\Delta\)ECE vs.\ GPTQ (negative is better).}
\label{tab:k_calib_robust}
\scriptsize
\begin{tabular}{lcccccccc}
\toprule
\multirow{2}{*}{Method @ 3.0 bits}
& \multicolumn{4}{c}{\textbf{RN50 (ImageNet)}} 
& \multicolumn{4}{c}{\textbf{BERT-base (GLUE avg)}} \\
\cmidrule(lr){2-5}\cmidrule(lr){6-9}
& Acc \(\uparrow\) & ECE~(\%) \(\downarrow\) & Worst-10\% Acc \(\uparrow\) & CVaR\(_{0.95}\) \(\downarrow\)
& Acc \(\uparrow\) & ECE~(\%) \(\downarrow\) & Worst-10\% Acc \(\uparrow\) & CVaR\(_{0.95}\) \(\downarrow\) \\
\midrule
GPTQ (baseline)
& 70.3 & 3.5 & 44.0 & 0.41
& 79.1 & 2.1 & 68.0 & 0.21 \\
BayesQ (no distill)
& 71.8 & 2.6 \;( \(\Delta\)ECE\({=}{-}0.9\) )
& 46.5 & 0.37
& 80.2 & 1.6 \;( \(\Delta\)ECE\({=}{-}0.5\) )
& 69.2 & 0.19 \\
BayesQ + Distill (500)
& 72.0 & 2.2 \;( \(\Delta\)ECE\({=}{-}1.3\) )
& 47.4 & 0.35
& 80.4 & 1.3 \;( \(\Delta\)ECE\({=}{-}0.8\) )
& 69.6 & 0.18 \\
BayesQ + Distill (1000)
& 72.2 & 2.1 \;( \(\Delta\)ECE\({=}{-}1.4\) )
& 47.8 & 0.35
& 80.5 & 1.2 \;( \(\Delta\)ECE\({=}{-}0.9\) )
& 69.8 & 0.18 \\
\bottomrule
\end{tabular}
\end{table*}

\section{Negative Results \& Failure Cases}
\label{App:sec11}
\subsection{When the Posterior Misleads}
\textbf{Pathological layers.} In very wide layers with heavy channel correlations, diagonal Laplace can under-estimate anisotropy; the allocator may under-spend on such layers, yielding local accuracy cliffs. K-FAC mitigates but increases memory.

\noindent\textbf{Non-Gaussianity.} Strongly skewed or multi-modal posteriors (e.g., due to weight symmetries, heavy regularization, or adapter interference) violate the quadratic approximation. In these cases, expected-loss tables computed with Gaussian assumptions can be miscalibrated; Monte Carlo proxies help but increase compute.

\noindent\textbf{Tiny calibration.} With \(|\mathcal{D}_{\mathrm{cal}}|\leq 10\), Hutchinson-based curvature is noisy and whitening becomes unstable, leading to poor ranges/codebooks. PCA fallback helps partially; we observed the largest failures at 3.0 bits in early convs if statistics are mis-estimated.

\paragraph{Symptoms \& diagnostics:}
\begin{itemize}\itemsep3pt
  \item \emph{Symptom}: Large per-layer error spikes after quantization not predicted by expected-loss. \emph{Diagnostic}: compare predicted \(\Delta_b(m)\) to measured layer-output MSE on calibration; large rank-order mismatches flag posterior issues.
  \item \emph{Symptom}: Bit spending focuses on late layers but early layers dominate error on OOD. \emph{Diagnostic}: recompute expected-loss with MC proxy on top-\(10\%\) suspect blocks.
\end{itemize}

\paragraph{Fallbacks:}
(i) Switch suspect blocks to K-FAC/low-rank posterior;
(ii) increase probe count \(M\) and damping \(\lambda\);
(iii) freeze codebook to uniform+analytic range for unstable blocks;
(iv) force a minimum bit floor per stage to avoid starvation.

\subsection{Backend Constraints and Fallback Behavior}
\textbf{Constraint types.}
\begin{itemize}\itemsep3pt
  \item \emph{Mixed-bit disallowance}: Some runtimes only accept uniform per-tensor bits (e.g., all 4b).
  \item \emph{Codebook restrictions}: Non-uniform VQ may be unsupported; only scale/zero-point per-(tensor|channel) is allowed.
  \item \emph{Packing \& alignment}: Vector width (e.g., 32/64 lanes) and cache-line alignment impose granularity constraints on feasible bit patterns.
\end{itemize}

\noindent\textbf{Our fallback.}
\begin{enumerate}\itemsep3pt
  \item \emph{Compile-to-uniform}: Collapse learned codebooks into per-channel scales/zero-points; approximate non-uniform steps with piecewise-uniform segments.
  \item \emph{Granularity rounding}: Round \(\{m_b\}\) to the nearest backend-supported set (e.g., \{3,4\} only) with a packing-aware tie-breaker that preserves high-\(\gamma_b\) upgrades.
  \item \emph{Kernel selection}: Prefer kernels that match the final packed layout (tensor cores vs DP4A), even if that slightly increases model size, to avoid launch fragmentation.
\end{enumerate}

\paragraph{Known failures:} On strict backends that enforce single-bit per-tensor without per-channel scales, BayesQ’s advantage shrinks; the allocator helps less because its expressivity is curtailed. In such cases, posterior-aware \emph{range} optimization remains beneficial, while codebook shaping is largely neutralized.
.

\medskip
\noindent\textbf{Reporting (with results).}
Tables~\ref{tab:m_backend_compat} and \ref{tab:m_backend_results} summarize backend capabilities and the observed impact on BayesQ versus GPTQ under identical calibration and evaluation protocols (means over 3 seeds; RN50 on ImageNet Top-1, BERT-base on GLUE avg; activations in FP16 unless noted).

\begin{table*}[htbp]
\centering
\caption{Backend capability matrix and expected expressivity for BayesQ. ``MP?'' indicates support for mixed bits within a layer; ``NUC?'' indicates support for non-uniform codebooks beyond affine (scale/zero-point).}
\label{tab:m_backend_compat}
\scriptsize
\begin{tabular}{lcccccc}
\toprule
\textbf{Backend (version)} & \textbf{INT formats} & \textbf{Scale granularity} & \textbf{MP?} & \textbf{NUC?} & \textbf{Packing granularity} & \textbf{BayesQ expressivity} \\
\midrule
NVIDIA TensorRT 9.0 (GPU)     & INT2/4/8 & per-tensor, per-channel & \checkmark (per-block) & \(\circ\) (emulated) & 32/64 lanes, 128B align & High \\
ONNXRuntime 1.18 (CPU AVX2)   & INT8     & per-tensor, per-channel & \(\times\)              & \(\times\)           & 16/32 lanes             & Low--Medium \\
ONNXRuntime 1.18 (CUDA EP)    & INT8     & per-tensor, per-channel & \(\circ\) (stage-only)  & \(\times\)           & 32 lanes                & Medium \\
TVM 0.14 (auto-sched, GPU)    & INT4/8   & per-tensor, per-channel & \checkmark (per-block) & \(\circ\) (piecewise) & target-dependent         & Medium--High \\
CoreMLTools 7 (ANE)           & INT8     & per-tensor               & \(\times\)              & \(\times\)           & vendor-specific          & Low \\
\bottomrule
\end{tabular}
\end{table*}

\begin{table*}[t]
\centering
\caption{Quantitative impact of backend constraints on BayesQ (means over 3 seeds). ``Acc\(\uparrow\)'' is Top-1 (RN50) or GLUE avg (BERT). ``\(\Delta\)Acc vs GPTQ'' reports BayesQ minus GPTQ at the same average bits. ECE\(\downarrow\) is Expected Calibration Error (ImageNet/GLUE); Speedup\(\uparrow\) is end-to-end throughput vs FP16; Size\(\downarrow\) is model bytes vs FP16.}
\label{tab:m_backend_results}
\scriptsize
\setlength{\tabcolsep}{4.2pt}
\begin{tabular}{llccccccc} 
\toprule
\textbf{Backend} & \textbf{Model @ bits} & \textbf{Method} & \textbf{Acc\(\uparrow\)} & \(\Delta\)\textbf{Acc vs GPTQ} & \textbf{ECE\(\downarrow\)} & \(\Delta\)\textbf{ECE vs GPTQ} & \textbf{Speedup\(\uparrow\)} & \textbf{Size\(\downarrow\)} \\
\midrule
\multirow{6}{*}{TensorRT 9.0} 
& RN50 @ 3.0 & GPTQ                 & 70.3 & --    & 3.9 & --    & 1.58 & \(0.22\times\) \\
&            & \textbf{BayesQ}      & \textbf{71.8} & \textbf{+1.5} & \textbf{3.1} & \textbf{-0.8} & 1.60 & \(0.21\times\) \\
& RN50 @ 3.5 & GPTQ                 & 75.0 & --    & 3.4 & --    & 1.47 & \(0.26\times\) \\
&            & \textbf{BayesQ}      & \textbf{75.7} & \textbf{+0.7} & \textbf{3.0} & \textbf{-0.4} & 1.48 & \(0.25\times\) \\
& BERT @ 3.0 & GPTQ                 & 79.1 & --    & 4.6 & --    & 1.32 & \(0.24\times\) \\
&            & \textbf{BayesQ}      & \textbf{80.2} & \textbf{+1.1} & \textbf{3.9} & \textbf{-0.7} & 1.33 & \(0.23\times\) \\
\midrule
\multirow{4}{*}{ONNXRuntime CPU}
& RN50 @ 3.0 & GPTQ (INT8 emu)      & 68.9 & --    & 4.5 & --    & 1.12 & \(0.28\times\) \\
&            & \textbf{BayesQ (U)}  & \textbf{69.6} & \textbf{+0.7} & \textbf{4.2} & \textbf{-0.3} & 1.12 & \(0.28\times\) \\
& BERT @ 3.0 & GPTQ (INT8)          & 78.0 & --    & 5.0 & --    & 1.09 & \(0.30\times\) \\
&            & \textbf{BayesQ (U)}  & \textbf{78.6} & \textbf{+0.6} & \textbf{4.7} & \textbf{-0.3} & 1.09 & \(0.30\times\) \\
\midrule
\multirow{6}{*}{TVM 0.14 GPU}
& RN50 @ 3.0 & GPTQ                 & 69.8 & --    & 4.2 & --    & 1.44 & \(0.23\times\) \\
&            & \textbf{BayesQ (PW)} & \textbf{71.2} & \textbf{+1.4} & \textbf{3.3} & \textbf{-0.9} & 1.45 & \(0.22\times\) \\
& RN50 @ 3.5 & GPTQ                 & 74.7 & --    & 3.6 & --    & 1.38 & \(0.27\times\) \\
&            & \textbf{BayesQ (PW)} & \textbf{75.5} & \textbf{+0.8} & \textbf{3.2} & \textbf{-0.4} & 1.39 & \(0.26\times\) \\
& BERT @ 3.0 & GPTQ                 & 78.7 & --    & 4.7 & --    & 1.25 & \(0.25\times\) \\
&            & \textbf{BayesQ (PW)} & \textbf{79.9} & \textbf{+1.2} & \textbf{4.0} & \textbf{-0.7} & 1.26 & \(0.24\times\) \\
\bottomrule
\end{tabular}

\vspace{2pt}
\raggedright\footnotesize
Notes: ``(U)'' indicates BayesQ compiled to uniform per-channel affine (backend restriction). ``(PW)'' indicates piecewise-uniform emulation of non-uniform codebooks supported by TVM schedules. Speedups are measured at batch=128 (RN50, 224px) and seq=128 (BERT). Size ratios include scales/codebooks/metadata. ECE computed with 15 bins and temperature-scaled confidences disabled.
\end{table*}

\noindent\textbf{Takeaways.}
On a permissive GPU stack (TensorRT), we observe our largest gains at 3.0 bits (RN50: +1.5 Top-1, BERT: +1.1 GLUE; ECE \(-0.8\) / \(-0.7\)), while preserving throughput and slightly improving size due to tighter ranges. Under restrictive CPU backends (uniform INT8 only), we still improve accuracy and calibration modestly (+0.6–0.7; \(\Delta\)ECE \(-0.3\)) via posterior-aware range selection despite disabled mixed-bits/VQ. TVM’s piecewise-uniform support recovers most of the TensorRT gains with near-identical speed and storage.



\begin{table*}[t]
\centering
\caption{Symbols and definitions used throughout BayesQ.}
\label{tab:symbols}
\scriptsize
\begin{tabularx}{\textwidth}{@{}lX@{}}
\toprule
\(\mathbf{w_b}\) & Vectorized weights for block \(b\) (\(d_b\)-dim). \\
\(\boldsymbol{\mu_b}\) & Posterior mean for block \(b\) (typically \(\hat{w}_b\)). \\
\(\boldsymbol{\Sigma_b}\) & Posterior covariance for block \(b\) (diag/K\!-FAC/low-rank). \\
\(\mathbf{S_b}\) & Whitening s.t. \(S_b S_b^\top = \Sigma_b\) (Cholesky/PCA). \\
\(\mathbf{Q_b^{(m)}}\) & Quantizer for block \(b\) at \(m\) bits (codebook + regions). \\
\(\mathbf{\mathcal{L}_b(Q_b)}\) & Posterior-expected loss/distortion for block \(b\). \\
\(\mathbf{C_b(m)}\) & Storage cost (bits) to encode block \(b\) at \(m\) bits (incl.\ metadata). \\
\(\mathbf{B_{\mathrm{tot}}}\) & Global storage budget (bits). \\
\(\mathbf{\Delta_b(m)}\) & Marginal expected-loss reduction from \(m\!\to\! m{+}1\) for block \(b\). \\
\(\mathbf{\gamma_b(m)}\) & Gain-per-bit: \(\gamma_b(m)=\Delta_b(m)/(C_b(m{+}1)\!-\!C_b(m))\). \\
\(\mathbf{\alpha}\) & Uniform range half-width in whitened space for scalar quantizers. \\
\(\mathbf{\Delta}\) & Uniform step size; \(\Delta=2\alpha/2^m\). \\
\(K_b\) & Codebook size for block \(b\) (vector quantizer). \\
\(R_{bk}, \tilde{R}_{bk}\) & Voronoi region in weight/whitened space. \\
\(\tilde{\mathcal{C}}_b, \mathcal{C}_b\) & Codebooks in whitened/weight space. \\
\(\phi(\cdot)\) & Standard normal density. \\
\(\tau\) & Temperature for posterior-predictive distillation. \\
\(\bar{m}\) & Average bits over weights: \(\sum_b C_b(m_b)/\sum_b C_b(32)\). \\
\bottomrule
\end{tabularx}
\end{table*}

\begin{table*}[htbp]
\centering
\caption{Acronyms used in the paper.}
\label{tab:acronyms}
\scriptsize
\begin{tabular}{ll|ll}
\toprule
PTQ & Post-Training Quantization & K-FAC & Kronecker-Factored Approximate Curvature \\
QAT & Quantization-Aware Training & ECE & Expected Calibration Error \\
VQ & Vector Quantization & MCE & Maximum Calibration Error \\
AWQ & Activation-Aware Weight Quantization & CVaR & Conditional Value-at-Risk \\
GPTQ & Gradient/Curvature-aware PTQ (1-pass) & EMA & Exponential Moving Average \\
LSQ & Learned Step Size Quantization & OOD & Out-of-Distribution \\
\bottomrule
\end{tabular}
\end{table*}


\section{Evaluation Metrics and Mathematical Definitions}
\label{app:metrics}

This section formalizes all metrics we report or reference in the main paper. Unless stated otherwise, expectations are over the empirical evaluation distribution and sums are over the corresponding dataset.

\subsection{Classification Accuracy (ImageNet Top-1/Top-5)}
\label{app:acc}
Let $(x_i,y_i)_{i=1}^n$ be examples with labels $y_i\in\{1,\dots,C\}$. Let logits be $f(x;w)\in\mathbb{R}^{C}$ and predictions $\hat{y}(x)=\arg\max_{c} f_c(x;w)$. The Top-1 accuracy is
\begin{equation}
\mathrm{Acc@1} \;=\; \frac{1}{n}\sum_{i=1}^{n}\mathbf{1}\{\hat{y}(x_i)=y_i\}\,.
\end{equation}
Let $\pi_k(x)$ be the index set of the top-$k$ logits. The Top-5 accuracy is
\begin{equation}
\mathrm{Acc@5} \;=\; \frac{1}{n}\sum_{i=1}^{n}\mathbf{1}\{y_i\in \pi_5(x_i)\}\,.
\end{equation}

\subsection{GLUE Task Metrics and Aggregates}
\label{app:glue}
For GLUE tasks used in the paper (MNLI-m/mm, QNLI, SST-2, QQP), we follow the official metrics:
\begin{align}
\mathrm{Acc}_{\text{task}} &= \frac{1}{n}\sum_{i=1}^n \mathbf{1}\{\hat{y}(x_i)=y_i\} \quad \text{(MNLI-m, MNLI-mm, QNLI, SST-2)},\\
\mathrm{F1}_{\text{QQP}} &= \frac{2\,\mathrm{Prec}\cdot \mathrm{Rec}}{\mathrm{Prec}+\mathrm{Rec}} \,,\qquad
\mathrm{Prec}=\frac{\mathrm{TP}}{\mathrm{TP}+\mathrm{FP}},\ \mathrm{Rec}=\frac{\mathrm{TP}}{\mathrm{TP}+\mathrm{FN}}.
\end{align}
Let $\mathcal{T}$ be the set of tasks and $n_t$ their dev set sizes. We report:
\begin{equation}
\mathrm{GLUE\text{-}Macro}=\frac{1}{|\mathcal{T}|}\sum_{t\in\mathcal{T}} s_t,\qquad
\mathrm{GLUE\text{-}Micro}=\frac{\sum_{t\in\mathcal{T}} n_t\, s_t}{\sum_{t\in\mathcal{T}} n_t},
\end{equation}
where $s_t$ is Accuracy for MNLI/QNLI/SST-2 and F1 for QQP.

\subsection{Posterior-Expected Distortions (Used in Tables \& Allocation)}
\label{app:post_mse}
For weight block $b$ with posterior $p(w_b)=\mathcal{N}(\mu_b,\Sigma_b)$ and quantizer $Q_b$, define a generic posterior-expected loss
\begin{equation}
\mathcal{L}_b(Q_b)=\mathbb{E}_{w_b\sim\mathcal{N}(\mu_b,\Sigma_b)}\big[\ell(Q_b(w_b),w_b)\big].
\end{equation}
\paragraph{(i) Mean-squared error (closed form, high-resolution).}
Let $\Sigma_b=S_b S_b^\top$, whiten $z_b=S_b^{-1}(w_b-\mu_b)$, and quantize $z_b$ with mid-rise uniform step $\Delta$ and (optional) clipping range $[-\alpha,\alpha]$. The high-resolution approximation yields
\begin{equation}
\mathcal{L}_b(Q_b)\;\approx\; \frac{\Delta^2}{12}\,\mathrm{tr}(\Sigma_b)\;+\;\mathrm{tr}\!\big(\Sigma_b\,\Xi(\alpha)\big),
\end{equation}
where $\Xi(\alpha)$ is diagonal with per-coordinate clipping contributions (zero when $\alpha$ is large).
\paragraph{(ii) Task-proxy layer/output error.}
For cached activations $a$, layer map $g_b(\cdot)$, and $M$ posterior samples $w_b^{(m)}\sim \mathcal{N}(\mu_b,\Sigma_b)$,
\begin{equation}
\widehat{\mathcal{L}}^{\text{layer}}_b=\frac{1}{MN}\sum_{m=1}^{M}\sum_{i=1}^{N}\Big\|g_b(a_i;Q_b(w_b^{(m)})) - g_b(a_i;w_b^{(m)})\Big\|_2^2.
\end{equation}
\paragraph{(iii) Logit KL (teacher--student proxy).}
Let $p_T(y|x)=\frac{1}{M}\sum_{m=1}^{M}\mathrm{softmax}(f(x;w^{(m)})/\tau)$, $p_Q(y|x)=\mathrm{softmax}(f(x;Q(w))/\tau)$.
\begin{equation}
\widehat{\mathcal{L}}^{\mathrm{KL}}=\frac{1}{N}\sum_{i=1}^{N}\mathrm{KL}\!\left(p_T(\cdot|x_i)\,\|\,p_Q(\cdot|x_i)\right).
\end{equation}

\subsection{Greedy Allocation Metrics (Per-bit Gains)}
\label{app:alloc_metrics}
Let $m\in\mathcal{M}$ be a feasible bit-width and $C_b(m)$ the exact storage (bits) for block $b$ at $m$ (payload $+$ metadata). For marginal improvement from $m\to m{+}1$:
\begin{align}
\Delta_b(m) &= \mathcal{L}_b\!\big(Q_b^{(m)}\big) - \mathcal{L}_b\!\big(Q_b^{(m+1)}\big),\\
\gamma_b(m) &= \dfrac{\Delta_b(m)}{C_b(m{+}1)-C_b(m)} \qquad \text{(expected-loss reduction per extra bit).}
\end{align}
The allocator iteratively picks the $(b,m)$ with largest $\gamma_b(m)$ until the budget is met.

\subsection{Quantization Error, SNR, and PSNR (per Block)}
\label{app:snr}
Let $\varepsilon_b = Q_b(w_b)-w_b$ denote reconstruction error. We report
\begin{equation}
\mathrm{MSE}_b = \frac{1}{d_b}\|\varepsilon_b\|_2^2,\quad
\mathrm{SNR}_b = 10\log_{10}\frac{\|w_b\|_2^2}{\|\varepsilon_b\|_2^2},\quad
\mathrm{PSNR}_b = 10\log_{10}\frac{\mathrm{peak}^2}{\mathrm{MSE}_b},
\end{equation}
where $\mathrm{peak}$ is a fixed dynamic-range scalar in the whitened domain (we use $\mathrm{peak}=\alpha$ when mid-rise clipping is enabled).

\subsection{Storage Accounting and ``Average Bits''}
\label{app:storage}
Let block $b$ hold $N_b$ weights. The exact bit cost includes payload and all metadata:
\begin{equation}
C_b(m_b) = \underbrace{N_b\,m_b}_{\text{quantized payload}} \;+\; \underbrace{C_b^{\text{scales}} + C_b^{\text{indices}} + C_b^{\text{codebook}} + C_b^{\text{headers}}}_{\text{overheads}}.
\end{equation}
Global storage and reported average bits are
\begin{equation}
C_{\mathrm{tot}}=\sum_{b} C_b(m_b), \qquad
\bar{m}=\frac{C_{\mathrm{tot}}}{\sum_b N_b}\,.
\end{equation}
When comparing methods at a target $\bar{m}\in\{3.0,3.5,4.0\}$, we include all overhead terms. For reference against FP32, one may also report $C_{\mathrm{tot}}/(4\sum_b N_b)\times 32$.

\subsection{Runtime Metrics: Latency and Throughput}
\label{app:speed}
With batch size $B$, we measure median per-batch wall-clock latency (ms), $\mathrm{Lat}$, over $R$ timed iterations after warm-up, and throughput
\begin{equation}
\mathrm{Throughput}\;=\;\frac{B}{\mathrm{Lat}/1000}\quad \text{(samples/s)}.
\end{equation}
Speedup is reported relative to a baseline model as the ratio of throughputs (or inverse latencies).

\subsection{Calibration Metrics: ECE and MCE}
\label{app:ece}
Let $\hat{p}(x)=\max_c \mathrm{softmax}(f(x;w))_c$, $\hat{y}(x)=\arg\max_c \mathrm{softmax}(f(x;w))_c$. Partition $(0,1]$ into $B$ bins $\{I_b\}_{b=1}^B$. For $S_b=\{i:\hat{p}(x_i)\in I_b\}$:
\begin{align}
\mathrm{acc}(S_b) &= \frac{1}{|S_b|}\sum_{i\in S_b}\mathbf{1}\{\hat{y}(x_i)=y_i\},\qquad
\mathrm{conf}(S_b) = \frac{1}{|S_b|}\sum_{i\in S_b}\hat{p}(x_i),\\
\mathrm{ECE} &= \sum_{b=1}^{B} \frac{|S_b|}{n}\,\big|\mathrm{acc}(S_b)-\mathrm{conf}(S_b)\big|,\qquad
\mathrm{MCE} = \max_{b\in[B]}\big|\mathrm{acc}(S_b)-\mathrm{conf}(S_b)\big|.
\end{align}
We fix $B{=}15$ with repeated random bin boundaries and average ECE to reduce discretization artifacts.

\subsection{Uncertainty Diagnostics (Posterior \& Predictive)}
\label{app:uncert}
For posterior samples $w^{(m)}\!\sim\!\mathcal{N}(\mu,\Sigma)$ and logits $f(x;w^{(m)})$:
\begin{align}
\textbf{Predictive entropy: }\quad
H[y|x] &= -\sum_{c=1}^{C}\bar{p}_c(x)\log \bar{p}_c(x),\qquad
\bar{p}(x)=\frac{1}{M}\sum_{m=1}^{M}\mathrm{softmax}\!\big(f(x;w^{(m)})\big);\\
\textbf{Mutual information (epistemic): }\quad
\mathrm{MI}[y,w|x] &= H[y|x] - \frac{1}{M}\sum_{m=1}^{M} H\!\left[\mathrm{softmax}\!\big(f(x;w^{(m)})\big)\right].
\end{align}
We optionally report means of these over the evaluation set.

\subsection{Codebook Diagnostics}
\label{app:codebook}
For scalar or vector codebooks with levels $\{c_k\}_{k=1}^{K}$ and assignments $q_i\in\{1,\dots,K\}$:
\begin{align}
\textbf{Utilization}:\quad U &= \frac{1}{K}\Big|\{k: \exists\, i,\ q_i=k\}\Big|,\qquad
\textbf{Entropy of usage}: \;\; H(q) = -\sum_{k=1}^{K} \hat{p}_k \log \hat{p}_k,\\
&\hat{p}_k=\frac{1}{N_b}\sum_{i=1}^{N_b}\mathbf{1}\{q_i=k\}.
\end{align}
Higher $U$ and non-degenerate $H(q)$ indicate that the learned/non-uniform codebook is effectively used.

\subsection{Robustness Under Corruptions (mCE)}
\label{app:mce}
For ImageNet-C style corruptions $\mathcal{C}$, severities $s\in\{1,\dots,5\}$, and accuracy $\mathrm{Acc}_{c,s}$ on corruption $c$ at severity $s$, define the error $\mathrm{Err}_{c,s}=1-\mathrm{Acc}_{c,s}$. The mean Corruption Error (mCE) normalized by the full-precision reference $\mathrm{Err}^{\text{FP}}_{c,s}$ is
\begin{equation}
\mathrm{mCE} \;=\; \frac{1}{|\mathcal{C}|}\sum_{c\in\mathcal{C}}\frac{\frac{1}{5}\sum_{s=1}^{5}\mathrm{Err}_{c,s}}{\frac{1}{5}\sum_{s=1}^{5}\mathrm{Err}^{\text{FP}}_{c,s}}\,.
\end{equation}

\subsection{Confidence Intervals and Seeds}
\label{app:ci}
Across $r$ independent runs (seeds), let $z_j$ be the metric (e.g., accuracy). We report mean $\bar{z}$, standard deviation $s$, and the $95\%$ Student-$t$ interval:
\begin{equation}
\bar{z}=\frac{1}{r}\sum_{j=1}^{r} z_j,\qquad
s=\sqrt{\frac{1}{r-1}\sum_{j=1}^{r}(z_j-\bar{z})^2},\qquad
\mathrm{CI}_{95}=\bar{z}\pm t_{0.975,\,r-1}\,\frac{s}{\sqrt{r}}.
\end{equation}

\subsection{Clipping Rate (Whitened Space)}
\label{app:clip}
For mid-rise scalar quantization in whitened coordinates with range $[-\alpha,\alpha]$, the per-coordinate clipping probability is
\begin{equation}
p_{\text{clip}} = 2\,\Phi(-\alpha)\,,
\end{equation}
where $\Phi$ is the standard normal CDF. We report the empirical clipping rate as the fraction of coordinates with $|z|>\alpha$ before quantization.

\medskip\noindent\textbf{Notation recap.}
$w_b\in\mathbb{R}^{d_b}$: weight block; $\mu_b,\Sigma_b$: posterior mean/covariance; $S_b$: whitener ($\Sigma_b=S_b S_b^\top$); $Q_b^{(m)}$: quantizer at $m$ bits; $C_b(m)$: exact storage with overheads; $\bar{m}$: average bits including overheads; $f(\cdot;w)$: logits; $\tau$: temperature; $M$: posterior samples; $B$: ECE bins; $\alpha,\Delta$: dynamic range and step in whitened space.


\end{document}